\documentclass{article}

\usepackage{microtype}
\usepackage{graphicx}
\usepackage{caption}
\usepackage{subfigure}
\usepackage{scalerel}
\usepackage{amsmath}
\usepackage{color}
\usepackage{listings}
\usepackage{float}
\usepackage{subfigure}
\usepackage{booktabs} 
\usepackage{layouts}

\definecolor{dkgreen}{rgb}{0,0.6,0}
\definecolor{gray}{rgb}{0.5,0.5,0.5}
\definecolor{mauve}{rgb}{0.58,0,0.82}

\usepackage{hyperref}

\usepackage[accepted]{icml2018}

\icmltitlerunning{Universal Planning Networks}

\begin{document}

\setlength{\textfloatsep}{1\baselineskip}

\twocolumn[
\icmltitle{Universal Planning Networks}

\icmlsetsymbol{equal}{*}
\icmlsetsymbol{help}{+}

\begin{icmlauthorlist}

\icmlauthor{Aravind Srinivas}{ucb}
\icmlauthor{Allan Jabri}{ucb}
\icmlauthor{Pieter Abbeel}{ucb}
\icmlauthor{Sergey Levine}{ucb}
\icmlauthor{Chelsea Finn}{ucb}
\end{icmlauthorlist}

\icmlaffiliation{ucb}{UC Berkeley, Computer Science}

\icmlcorrespondingauthor{Aravind Srinivas}{aravind@cs.berkeley.edu}

\icmlkeywords{Machine Learning, ICML}

\vskip 0.3in
]

\printAffiliationsAndNotice{} 

\begin{abstract} 

A key challenge in complex visuomotor control is learning abstract representations that are effective for specifying goals, planning, and generalization. To this end, we introduce universal planning networks (UPN). UPNs embed differentiable planning within a goal-directed policy. This planning computation unrolls a forward model in a latent space and infers an optimal action plan through gradient descent trajectory optimization. The plan-by-gradient-descent process and its underlying representations are learned end-to-end to directly optimize a supervised imitation learning objective. We find that the representations learned are not only effective for goal-directed visual imitation via gradient-based trajectory optimization, but can also provide a metric for specifying goals using images.  The learned representations can be leveraged to specify distance-based rewards to reach new target states for model-free reinforcement learning, resulting in substantially more effective learning when solving new tasks described via image-based goals. We were able to achieve successful transfer of visuomotor planning strategies across robots with significantly different morphologies and actuation capabilities. 

\end{abstract}

\section{Introduction} 
\begin{figure}[ht]
\centering
\subfigure[Universal Planning Network (UPN)]{\label{fig:teaserarch}
\includegraphics[width=0.90\linewidth]{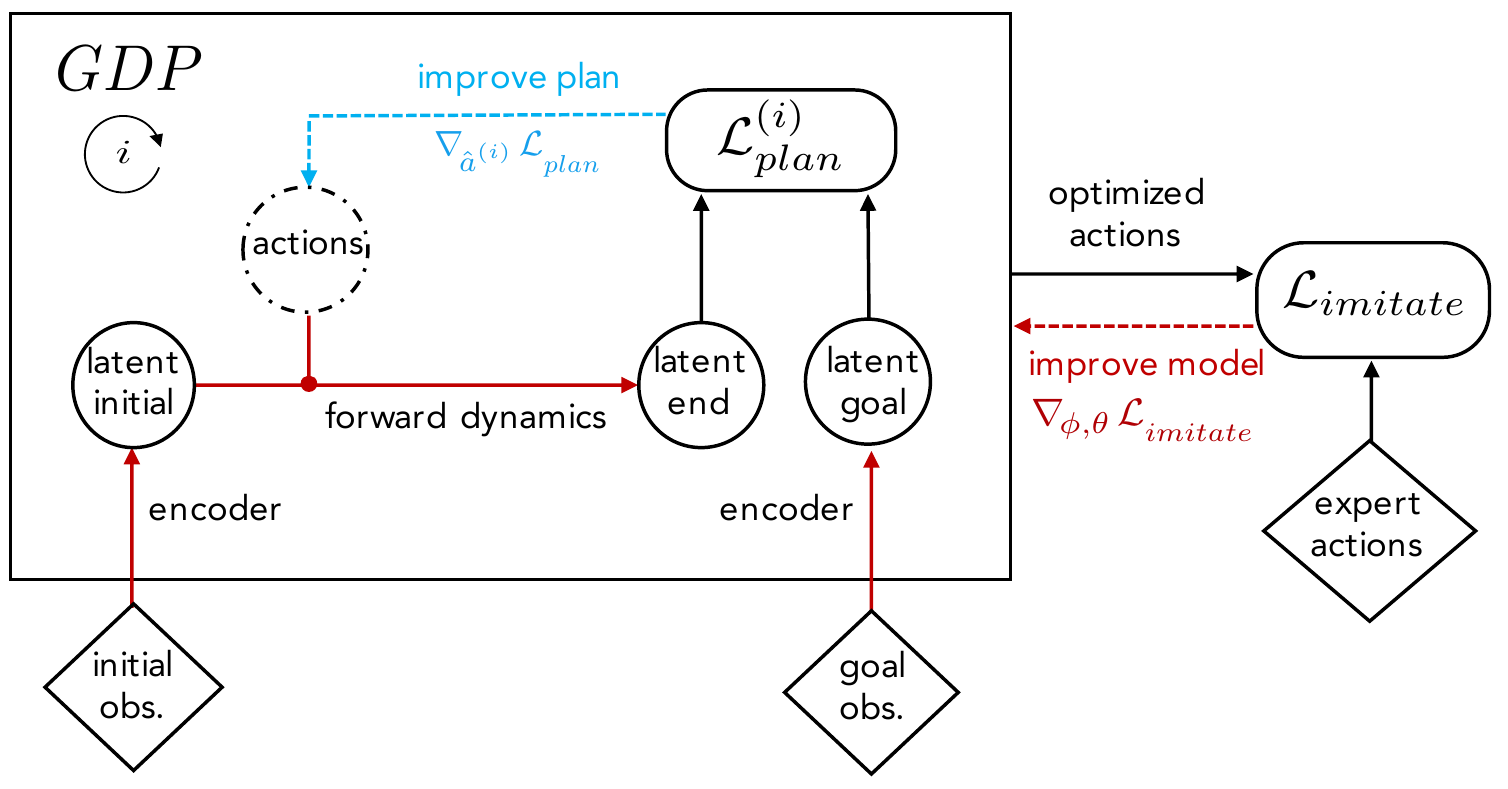}}
\vspace{-2mm}
\subfigure[Leveraging learned latent representations]{\label{fig:teasertransfer}
\includegraphics[width=0.88\linewidth]{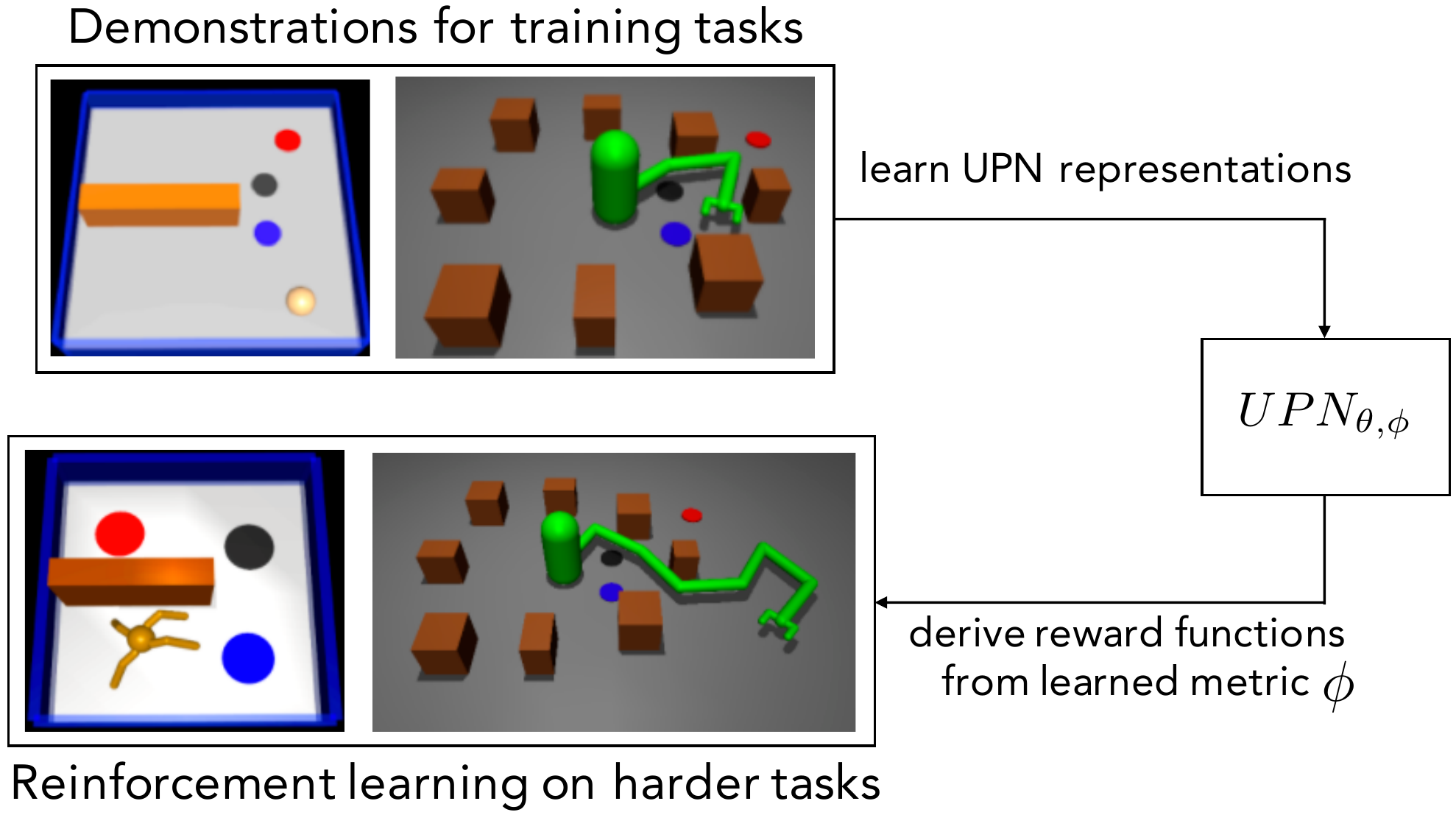}}

\vspace{-1mm}
    \caption{ An overview of the UPN, which embeds a gradient descent planner (GDP) in the action-selection process. We demonstrate transfer to different, harder control tasks, including morphological (yellow point robot to ant) and topological (3-link to 7-link reacher) variants, as shown above.} 
    \label{fig:teaser}
\end{figure}

Learning visuomotor policies is a central pursuit in building machines capable of performing complex skills in the variety of unstructured and dynamic environments seen in the real world \cite{levine2016end, pinto2016curious}. A key challenge in learning such policies lies in acquiring representations of the visual environment and its dynamics that are suitable for control. This challenge arises both in the construction of the policy itself and in the specification of the task. Extrinsic and perfect reward signals are typically not available for real world reinforcement learning and users must manually specify tasks via hand-crafted rewards with hand-crafted representations. To automate this process, some prior methods have proposed to specify tasks by providing an image of the goal scene \cite{deguchi1999image, watter2015embed, finn2016deep}. However, a reward that measures success based on matching the raw pixels of the goal image is far from ideal: such a reward is both uninformative and overconstrained, since matching all pixels is usually not required for succeeding in tasks. If we can automatically identify the right representation, we can both accelerate the policy learning process and  simplify the specification of tasks via goal images. Prior work in visual representation learning for planning and control has relied predominantly on unsupervised or self-supervised objectives~\cite{watter2015embed, finn2016deep}, which in principle only provide an indirect connection to the utility of the representation for the underlying control problem. Effective representation learning for planning and control remains an open problem.

In this work, instead of learning from unsupervised or auxiliary objectives and expecting that useful representations should emerge, we directly optimize for {\it plannable} representations: learning representations such that gradient-based planning
is successful with respect to the goal-directed task. To that end, we propose universal planning networks (UPN), a neural network architecture that can be trained to acquire a plannable representation. By embedding a differentiable planning computation inside the 
policy, our method enables joint training of the planner and its underlying latent encoder and forward dynamics representations. An outer imitation learning objective ensures that the learned representations are directly optimized for successful gradient-based planning on a set of training demonstrations. However, in principle, the architecture could also be trained with other policy optimization techniques such as those from reinforcement learning. An overview is provided in Figure \ref{fig:teaserarch}.

We demonstrate that the representations learned by UPN not only support gradient-based trajectory optimization for successful visual imitation, but in fact acquire a meaningful encoding of state, which can be used as a metric for task-specific latent distance to a goal. We find that we can reuse this representation to specify latent distance-based rewards to reach new target states via standard model-free reinforcement learning, resulting in substantially more effective learning when using image targets. These properties are naturally induced by the agent's reliance on the minimization of the latent distance between its predicted terminal state and goal state throughout the planning process. By learning {\it plannable} representations, the UPN learns an optimizable latent distance metric. Our findings are based on a new suite of challenging vision-based simulated robot control tasks that involve planning.

At a high-level, our approach is a goal-conditioned policy architecture that leverages a gradient-based planning computation in its action-selection process. While the architecture is agnostic to the objective function in the outer loop, we will focus on the imitation learning setting. From the perspective of representation learning, our method provides a way to learn more effective representations suitable for specifying perceptual reward functions, which can then be used, for example, with a model-free reinforcement learner. In terms of meta-learning, our architecture can be seen as learning a planning computation by learning representations that are in some sense {\it traversible} by gradient descent trajectory optimization for satisfying the outer meta-objective.

In extensive experiments, we show that (1) UPNs learn effective visual goal-directed policies more efficiently (that is, with less data) than traditional imitation learners; (2) the latent representations induced by optimizing for successful planning can be leveraged to transfer task-related semantics to other agents for more challenging tasks through goal-conditioned reward functions, which to our knowledge has previously not been demonstrated; and (3) the learned planning computation improves when allowed more updates at test-time, even in scenarios of less data, providing encouraging evidence of successful meta-learning for planning.

\section{Universal Planning Networks}

\begin{figure*}[]
\centering
\includegraphics[width=0.95\textwidth]{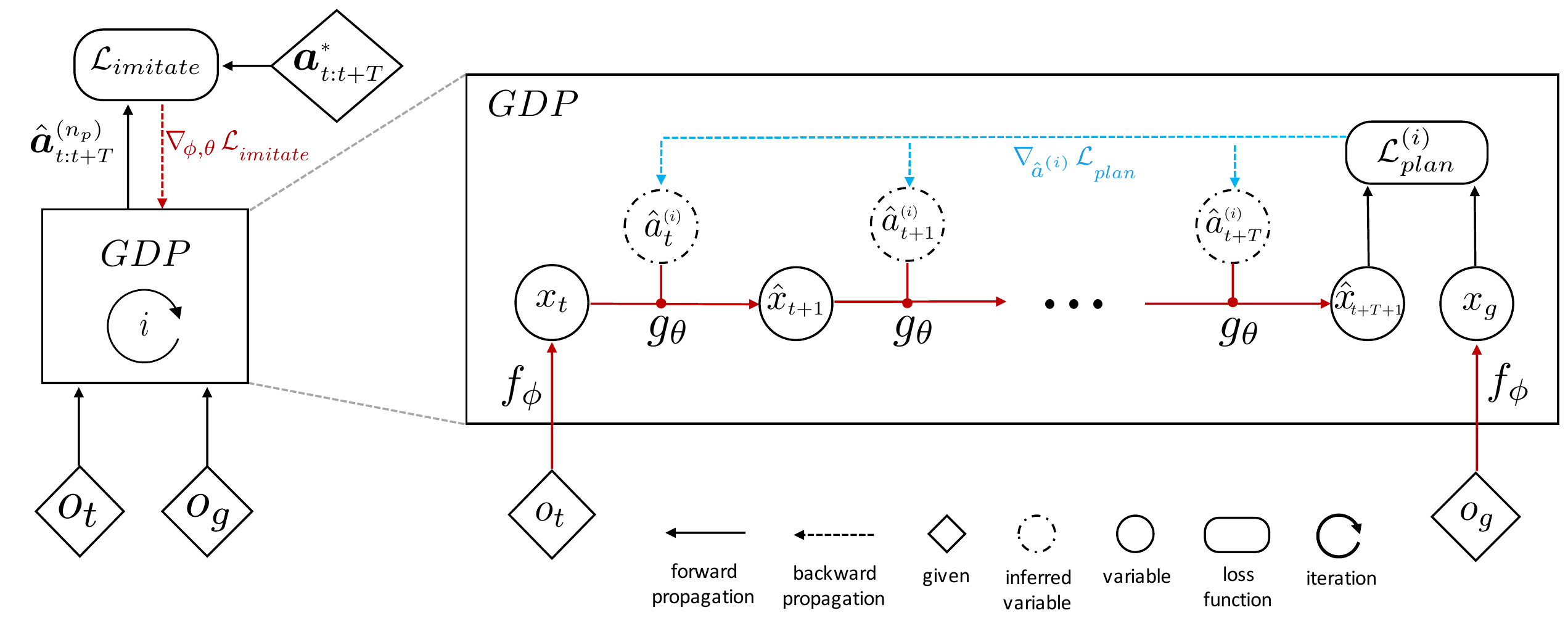}
\caption{An overview of the proposed method. Given an initial $o_t$ and a goal $o_g$, the GDP (\textit{gradient descent planner}) uses gradient descent to optimize a plan to reach the goal observation with a sequence of actions in a latent space represented by $f_\phi$. This planning process forms one large computation graph, chaining together the sub-graphs of each iteration of planning. The learning signal is derived from the (outer) imitation loss and the gradient is back-propagated through the entire planning computation graph. The blue lines represent the flow of gradients for planning, while the red lines depict the meta-optimization learning signal and the components of the architecture affected by it. Note that the GDP iteratively plans across $n_p$ updates, as indicated by the $i^{\textrm{th}}$ loop.}
\label{fig:mainfig}
\end{figure*}


Model-based approaches leverage forward models to search for, or \textit{plan}, sequences of actions to achieve goal states such that a planning objective is minimized. Forward modeling supports simulation of future state and hence, in principle, should allow for planning over extended horizons. In the absence of known environment dynamics, a forward model must be learned. 
Differentiable forward models allow for end-to-end training of model-based planners, as well as planning by back-propagating gradients with respect to input actions \cite{Schmidhuber90anon-line, henaff2017model}.

Nevertheless, learned forward models may: (1) suffer from function approximation modeling error, especially in complex, high-dimensional environments, (2) capture irrelevant details under the incentive to reduce model-bias, as is often the case when learning directly from pixels, and (3)  not necessarily align with the task and planning problem at hand, such that the inferred plans are sub-optimal even if the planning objective is optimized.

These issues motivate a central idea of the proposed method: instead of learning from surrogate unsupervised or auxiliary objectives, we directly optimize for what we care about, which is, representations with which  gradient-based trajectory optimization leads to the desired actions.
We study a model-based architecture that performs a differentiable planning computation in a latent space jointly learned with forward dynamics, trained end-to-end to encode what is necessary for solving tasks by gradient-based planning.

\subsection{Learning to Plan} \label{learningtoplan}


The UPN computation graph forms a goal-directed policy supported by an iterative planning algorithm.
Given initial and goal observations ($o_t$ and $o_g$) as input images, the model produces an optimal plan $\hat{a}_{t:t+T}$ to arrive at $o_g$, where $\hat{a}_t$ denotes the predicted action at time $t$. The computation graph consists of a pair of tied encoders that encode both $o_t$ and $o_g$, and their features are fed into a gradient descent planner (GDP), which produces the action $a_t$ as output. The GDP uses a neural network encoder and forward dynamics model to simulate transitions in a learned latent space and is thus fully differentiable. An overview of the method is presented in Figure \ref{fig:mainfig}.

The GDP uses gradient descent to optimize for a sequence of actions $\hat{a}_{t:t+T}$ to reach the encoded goal observation $o_g$ from an initial $o_t$. Since the model is differentiable, backpropagation through time allows for computing the gradient with respect to each planned action in order to end up closer to the desired goal state. Each iteration of the GDP thus involves unrolling the trajectory of latent state encodings using the current planned actions, and taking a step along the gradient to improve the planning objective. The cumulative planning process forms a large, differentiable computation graph, chaining together each iteration of planning.

The actual learning signal is derived from an outer loss function, which supervises the entire computation graph (including the GDP) to output the correct action sequence. The outer loss can in principle take any form, but in this work we use an imitation learning loss and supervise the entire model with demonstrations. The outer loss provides task-specific grounding to optimize for representations that support effective iterative planning for the task and environment at hand, as the gradient is back-propagated through the entire iterative planning computation graph.

Training thus involves nested objectives. One can view the learning process as first deriving a plan to achieve the goal and then updating the model parameters to make the planning procedure more effective for the outer objective. In other words, we seek to learn the planning computation through its underlying representations for latent state encoding and latent forward dynamics. 

{\textbf{Parameters:}} The model is composed of a forward dynamics model $g_\theta$ and an encoder $f_\phi$, where $\theta$ and $\phi$ are neural network parameters that are learned end-to-end:

\hspace{16mm}$x_t = f_\phi(o_t) \hspace{6mm} \hat{x}_{t+1} = g_\theta(x_t, a_t)$


Specifically, $f_\phi$ is a convolutional network and $g_\theta$ is a fully connected network.
Further architectural details can be found in the supplementary. \footnote{We note that one could also use an action encoder $h_\alpha(a_t) = u_t$, with $g_\theta$ operating on $x_t$ and $u_t$. A temporal encoder $h$ would allow for abstract sequences of actions (options), for an option conditioned latent forward model $g_\theta$. We work with flat sequences of actions, leaving hierarchical extensions for future work.}

{\textbf{Planning by Gradient Descent:}} The planner starts with an element-wise randomly initialized plan $\hat{a}_{t:t+T}^{(0)} \sim \mathcal{U}(-1, 1)$
and aims to minimize the distance between the predicted terminal latent state and the encoded goal observation. $T$ denotes the horizon over which the agent plans, which can depend on the task and hence may be treated as a hyper-parameter, while $n_{p}$ is the number of planning updates performed. Algorithm \ref{alg:innerloop} describes the iterative optimization procedure that is implemented by the GDP. 

\begin{algorithm}[h]
    \caption{$GDP(o_t, o_g, \alpha) \rightarrow \hat{a}_{t:t+T}$}
    \label{alg:innerloop}
\begin{algorithmic}
\REQUIRE $\alpha$: hyperparameter for step size 
\STATE Randomize an initial guess for the optimal plan $\hat{a}_{t:t+T}^{(0)}$
\FOR{$i$ from $0$ to $n_p-1$}
    \STATE Compute $x_t = f_\phi(o_t)$, $x_g = f_\phi(o_g)$
    \FOR{$j$ from $0$ to $T$}
     \STATE       $\hat{x}_{t+j+1}^{(i)} = g_\theta(\hat{x}_{t+j}^{(i)},{\hat{a}_{t+j}}^{(i)}) $
    \ENDFOR
    \STATE Compute $\mathcal{L}_{plan}^{(i)} = {||{\hat{x}_{t+T+1}}^{(i)} - x_g||_2}^2$ 
    \STATE Update plan: ${\hat{a}_{t:t+T}}^{(i+1)} = {\hat{a}_{t:t+T}}^{(i)} - \alpha \nabla_{{\hat{a}_{t:t+T}}^{(i)}} \mathcal{L}_{plan}^{(i)} $
\ENDFOR
    \STATE Return ${\hat{a}_{t:t+T}}^{(n_p)}$
\end{algorithmic}
\end{algorithm}

{\textbf{Huber Loss:}}
In practice, for $\mathcal{L}_{plan}^{(i)}$, we use a Huber Loss centered around $x_g$ for well-behaved inner loop gradients instead of a direct quadratic ${||{\hat{x}_{t+T+1}}^{(i)} - x_g||_2}^2$ . This usage is inspired from the Deep Q Networks paper of \citet{mnih2015human} and similar metrics have also been used by \citet{levine2016end} and \citet{sermanet2017time}.

{\textbf{Action selection at test-time:}} At test-time, Algorithm \ref{alg:innerloop} can be used to produce a sequence of actions. A more sophisticated approach is to use Algorithm \ref{alg:innerloop} to re-plan at each timestep. The agent first plans a trajectory suitable to reach $o_g$ from $o_t$, but only executes the first action, before replanning. This allows the agent to achieve goals requiring longer planning horizons at test-time even if the GDP was trained with a shorter horizon. This amounts to using model-predictive control (MPC) over our learned planner.

\subsection{Imitation as the Outer Objective}

An idea central to our approach is to directly optimize the planning computation for the task at hand, through the outer objective.  Though in this work we study the use of an imitation loss as the outer objective, the policy can in principle be trained through any gradient-based policy search method including policy gradients \cite{Schulman:EECS-2016-217} and value functions \cite{suttonbook}.

To learn parameters $\phi$ and $\theta$, we do not directly optimize the planning error under $\mathcal{L}_{plan}$, but instead learn the planner insofar as it can imitate an expert agent by iteratively applying $\mathcal{L}_{plan}$ (Algorithm \ref{alg:outerloop}). The model is therefore trained to plan in such a way as to produce actions that match the expert demonstrations.

Note that the subroutine $GDP(o_t, o_g, \alpha)$ is an accumulated computation graph composed of several iterations of planning,
each of which includes encoding observations and unrolling of latent forward dynamics through time.
Learning end-to-end thus requires that we back-propagate the behavior cloning loss under the produced plan through the GDP subroutine as depicted in Figure \ref{fig:mainfig}. We note that the gradients obtained on the network parameters $\theta$ and $\phi$ from the outer objective are composed of first-order derivatives of these parameters. Therefore, even though the computation graph of UPN may seem long and complicated, it is not prohibitively expensive to compute or difficult to optimize.

\begin{algorithm}[t]
   \caption{Learning the Planner via Imitation}
   \label{alg:outerloop}
\begin{algorithmic}
\REQUIRE $GDP(o_t, o_g, \alpha)$, expert $a^{*}_{t:t+T}$, step sizes $\alpha, \beta$

\FOR{$n$ from $1$ to $N$}
    \STATE Sample a batch of demonstrations $o_t, o_g, a^*_{t:t+T}$
    \STATE Compute $\hat{a}_{t:t+T} = GDP(o_t, o_g, \alpha)$
    \STATE Compute $\mathcal{L}_{imitate} = ||\hat{a}_{t:t+T} - a^{*}_{t:t+T}||^2_2$
    \STATE Update  $\theta := \theta - \beta \nabla_{\theta}\mathcal{L}_{imitate}$
    \STATE Update $\phi := \phi - \beta \nabla_{\phi}\mathcal{L}_{imitate}$
\ENDFOR

\end{algorithmic}
\end{algorithm}

In learning to plan via imitation, the agent jointly optimizes for latent state and dynamics representations that capture notions of state comparison useful for the imitation task and that are in some sense {\it traversible}  by gradient descent trajectory optimization. This is naturally induced by the agent's reliance on the minimization of the latent distance between its predicted terminal state and goal state throughout the planning process. Thus, in requiring  {\it plannable} representations, the encoder learns an optimizable latent distance metric. This is key to the viability of using the learned latent space as a metric from which to derive reward functions for reinforcement learning.

\subsection{Reinforcement Learning with a UPN Latent Space} \label{rlexplanation}

Reward functions are difficult to manually specify for visuomotor tasks described via image targets. Rewards purely based on pixel errors are meaningless, particularly when dealing with high dimensional images. A solution to this problem is to specify rewards in terms of distance to the target image in an abstract representation. There have been attempts in the past to learn such abstract representations. \citet{watter2015embed} and \citet{finn2016deep} take the unsupervised learning route using autoencoders, while \citet{sermanet2017time} attempt to fine-tune representations from Imagenet using auxiliary losses tailor-made for robotic manipulation. With UPN having been trained for acquiring {\it plannable} representations, it is only natural to expect that its latent space encoded by $f_\phi$  serves the role of an abstract representation where rewards can be specified for performing reinforcement learning on visuomotor tasks with image targets. More specifically, we can exploit the learned $f_\phi$ from UPN to provide reward functions of the form $r(o_t, o_g) = -||f_{\phi}(o_t) - f_{\phi}(o_g)||^2_2$. In practice, we use the Huber Loss around $o_g$ to stay consistent with the metric the UPN was trained with. Further, we had more success normalizing the distance metric to lie in the interval $[0,1]$ by passing the negative of the distance through an exponential. The details are highlighted in the supplementary.

Figure \ref{fig:rlfig} visually depicts the reinforcement learning process. While performing reinforcement learning on the new tasks, the agent gets access to its own embodiment $s_t$ (joint angles and velocities) and the feature vector of the goal $f_\phi(o_g)$ as its input observations. The agent optimizes for the perceptual rewards computed from UPN and does not receive any extrinsic rewards from the environment. Providing the feature vector of the goal is necessary when the evaluation success is averaged over multiple goals at test time. In case of a single fixed goal, the evaluation success is averaged over different initial configurations of the robot which can be captured in the information provided via $s_t$ and the goal feature vector becomes redundant. Unless specified, we evaluate using multiple goals at test-time and feed in $f_\phi(o_g)$ as an additional input to the reinforcement learning agent, thereby making the policy architecture at test time {\it universal}.

\begin{figure}[ht]
\centering

\includegraphics[width=0.35\textwidth]{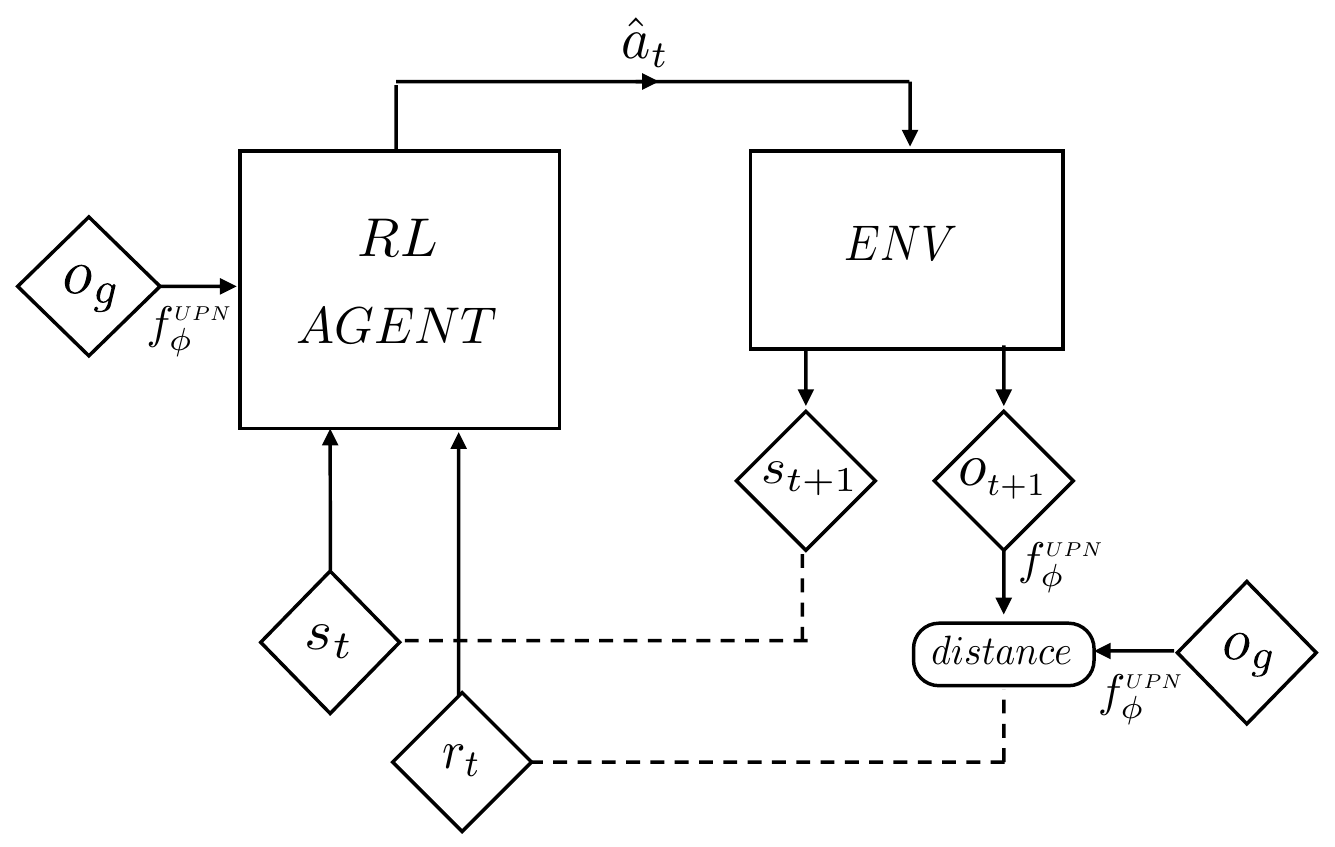}
\caption[width=0.7\textwidth]{A reinforcement learning agent can derive rewards from the latent representations learned by the UPN. The rewards are based on the difference between $o_t$ and $o_g$ in the abstract representation, while the policy is conditioned on joint angles and velocities specific to the agent, $s_t$; and the feature vector of the goal, $f_\phi^{UPN}(o_g)$. The agent has to reason about the goals and how to achieve them based on the learned features from UPN.}
\label{fig:rlfig}
\end{figure}

\section{Related Work}

\begin{figure*}[t]
\centering
\subfigure[Pointmass config. 1]{\label{fig:pointmassa}\includegraphics[width=0.22\linewidth]{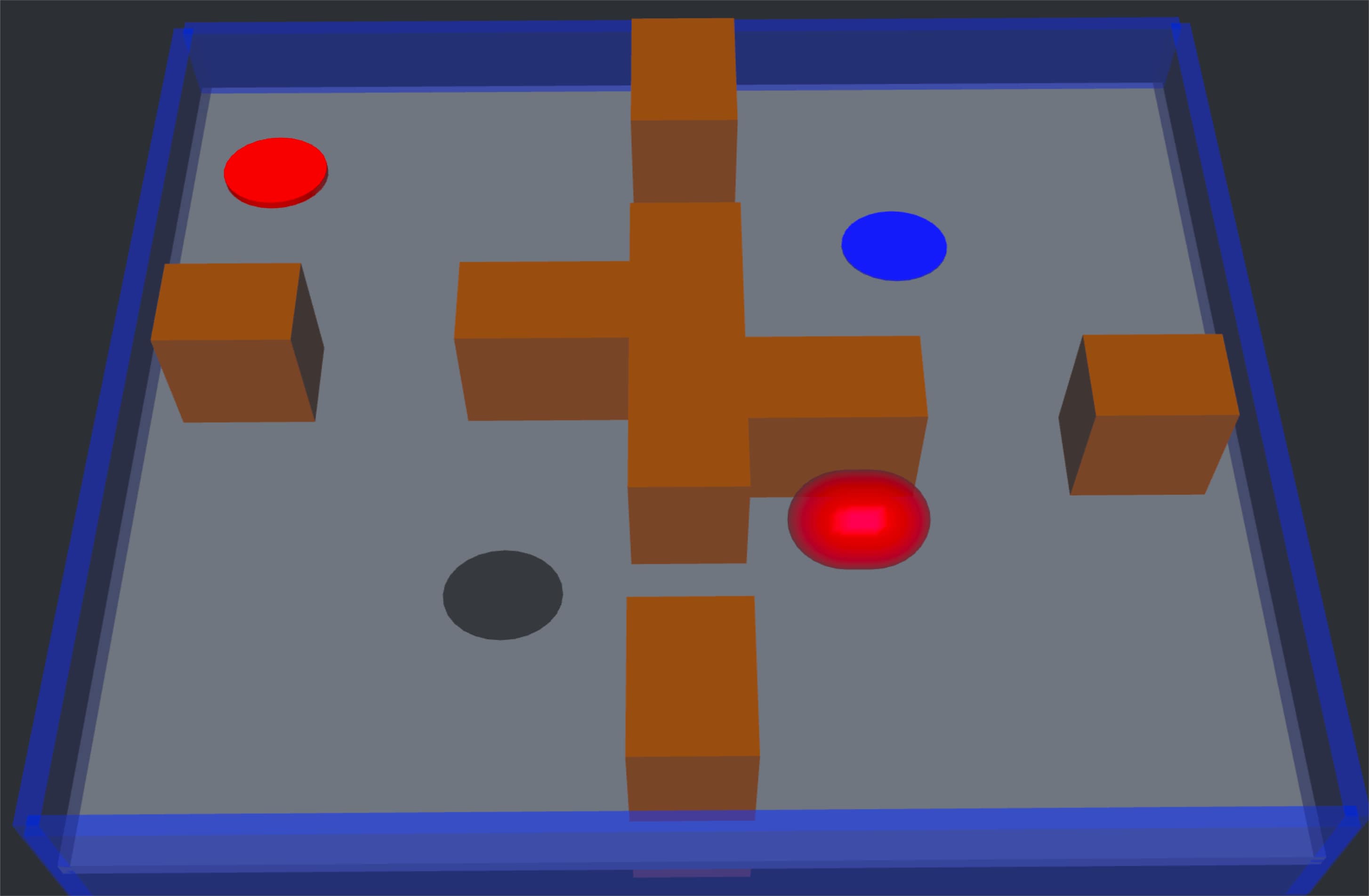}}
\hspace{1mm}    \subfigure[Pointmass config. 2]{\label{fig:pointmassb}\includegraphics[width=0.22\linewidth]{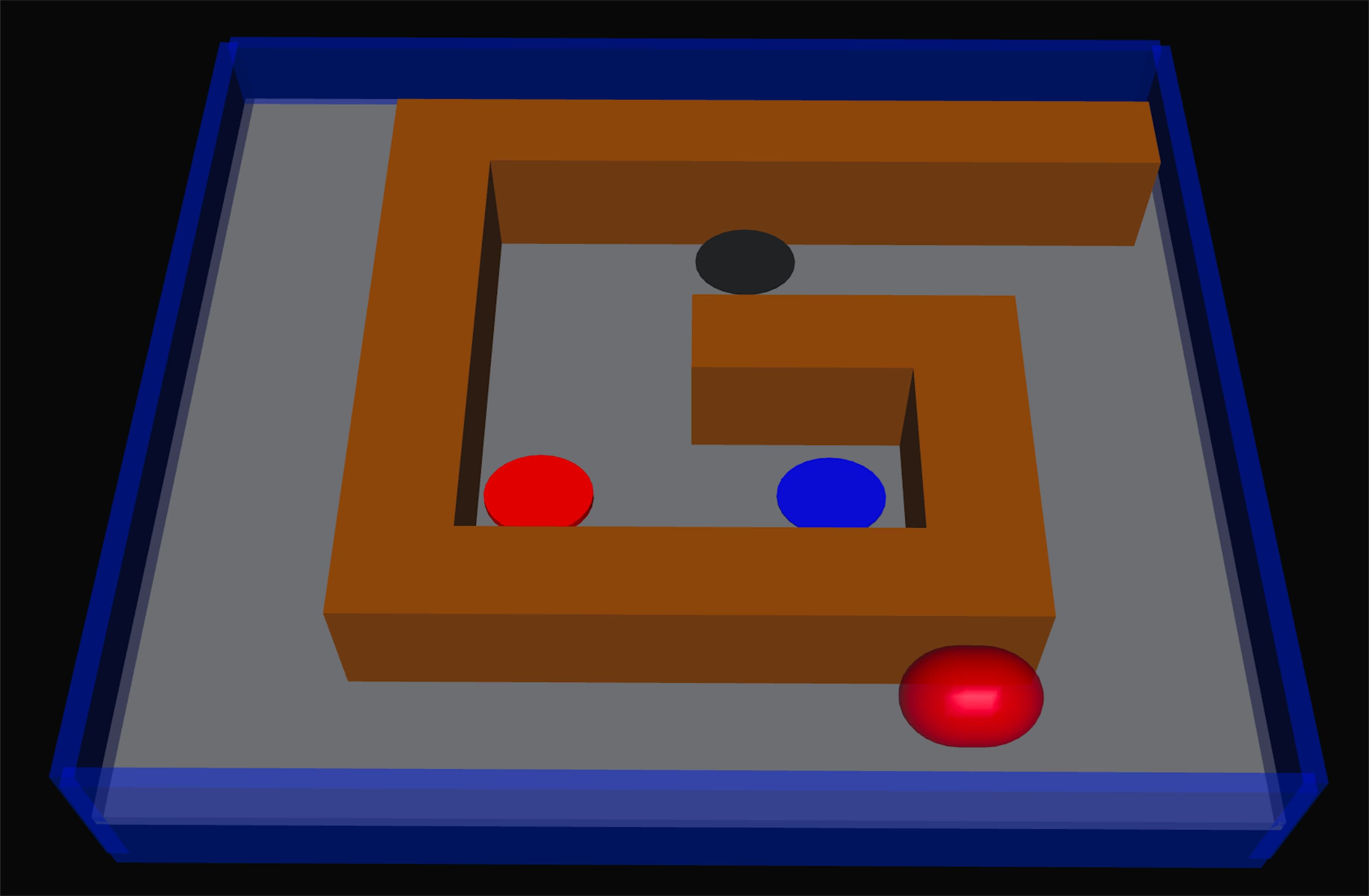}}
\hspace{1mm}   \subfigure[Reacher config. 1]{\label{fig:reachera}\includegraphics[width=0.22\linewidth]{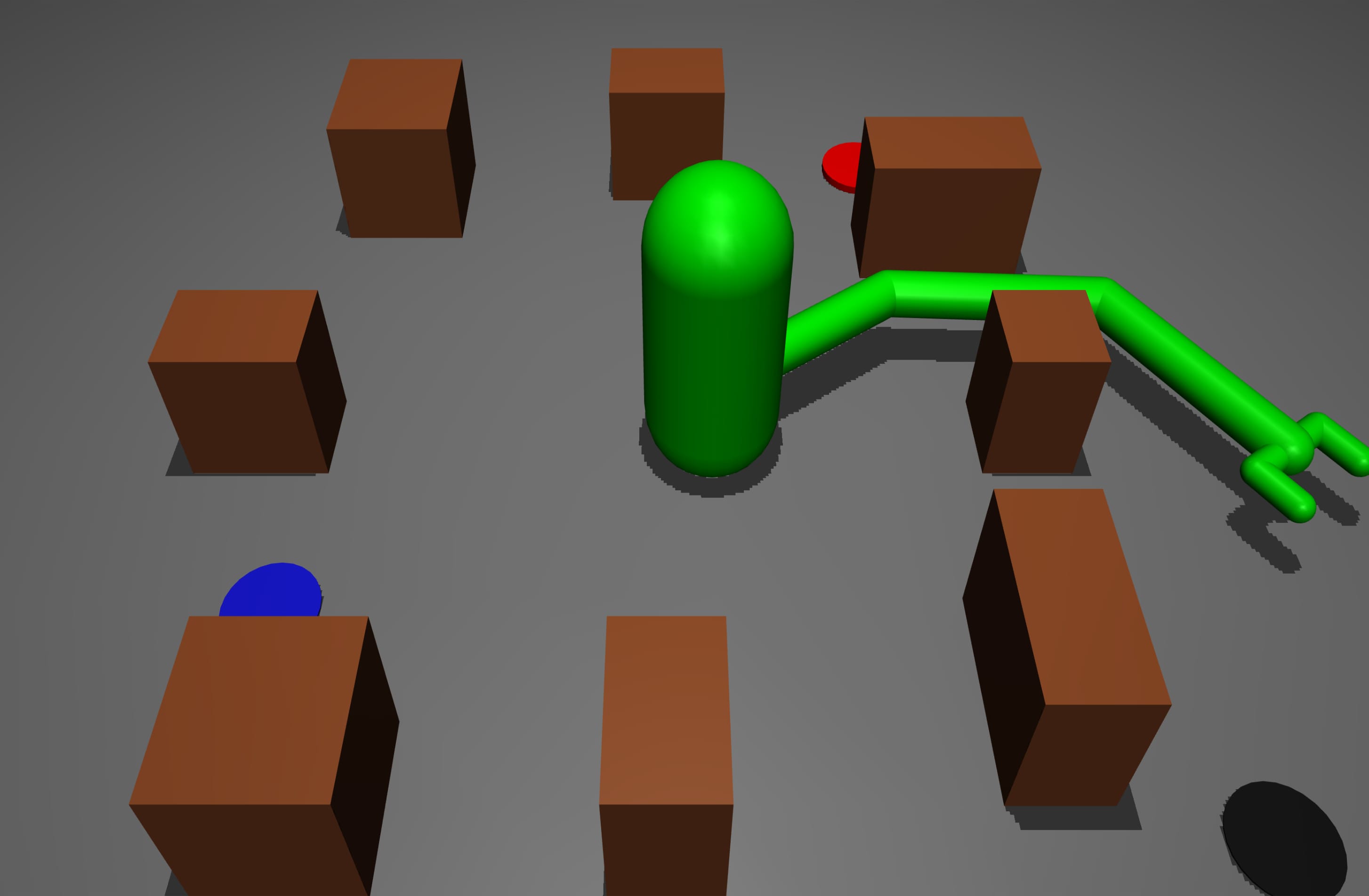}}
\hspace{1mm}  \subfigure[Reacher config. 2]{\label{fig:hardobstacle}\includegraphics[width=0.22\linewidth]{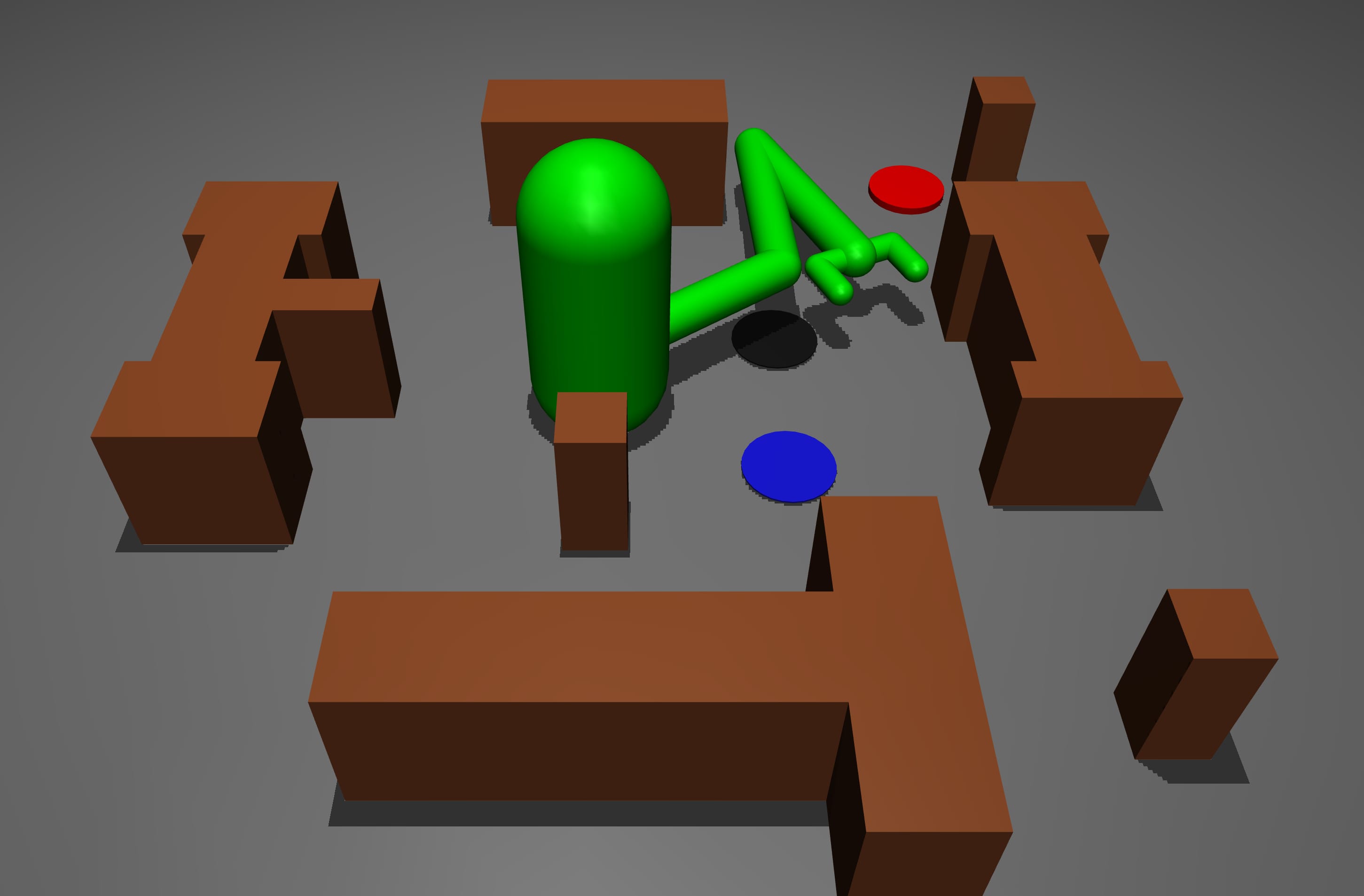}}
    \caption{Examples of the visuomotor tasks considered for the zero shot generalization study. We consider two 2D robot models: a force-controlled point robot and a 3-link torque-controlled reacher robot. We consider two types of generalization: fixing the obstacles while varying the target goals (FOVG) and varying both the obstacle and target goals (VOVG). These tasks require non-trivial generalization combining visual planning with low level motor control.}
    \label{fig:tasksnapshots}
\end{figure*}{}

Our work is primarily concerned with learning representations that can support planning for tasks described through an image target. \citet{watter2015embed} and \citet{finn2016deep} take an unsupervised learning approach to learning such representations, which they use for planning with respect to target images using iLQR \cite{tassa2012synthesis}. However, reconstructing all the pixels in the scene could lead to the encoding of state variables not necessarily useful in the context of planning \cite{higgins2017darla} and discard state variables that are not visually prominent (\citet{goodfellow2016deep}, Chapter 15). Our approach avoids this problem by explicitly optimizing a representation for {\it plannability through gradient descent} as the only criterion. Self-supervised methods that avoid pixel reconstruction by using other intermediate forms of supervision that can be obtained automatically from the data have also been used to learn representations for visuomotor control \cite{sermanet2016unsupervised, sermanet2017time}. We again differ by optimizing directly for what we need: plannable representations, instead of intermediate objectives. While the goal in \citet{sermanet2017time} is to recover a reward function to mimic specific demonstrations, our goal is to acquire a more broadly applicable representation from demonstrations that can then be used to perform \textit{new} tasks using just a single goal image.

There has been work in learning state representations usable for model-free RL when provided rewards \cite{lange2012autonomous, jonschkowski2015learning, jonschkowski2017pves, higgins2017darla, 8276247}.  The key difference in our work is that we focus on learning representations that can be used for defining metric-based rewards for new tasks, as opposed to just learning state representations for RL from external environment rewards.

Learning representations capable of providing distance metric based rewards naturally relates to inverse reinforcement learning (IRL) \cite{ng2000algorithms, abbeel2004apprenticeship,finn2016guided,ho2016generative, baram2017end} and reward shaping \cite{ng1999policy}. IRL methods attempt to learn a reward function from expert demonstrations which could then be used to optimize a traditional reinforcement learner. However, IRL from raw pixels is challenging due to the lack of sufficient constraints in the problem definition; only a couple of methods have successfully applied IRL to images, and to do so have relied on human domain knowledge~\cite{wulfmeier2016incorporating} and pre-training~\cite{li2017inferring}. Our work can be viewed as connecting IRL and reward shaping: learning representations {\it amenable} to gradient-based trajectory optimization is one way to extract a perceptual reward function. However, we differ significantly from conventional IRL in that our derived reward functions are effective even for new tasks.

From an architectural standpoint, we embed a differentiable planner within our computation graph. 
Value iteration networks of \citet{tamar2016value} embed an approximate differentiable value iteration computation, though their architecture only supports discrete planning and is evaluated on tasks with sparse state transition probabilities. We seek a more general planning computation for more complex transition dynamics and continuous actions suitable for motor control from raw pixels. \citet{tamar2017learning} attempt to learn an embedded differential MPC controller by reshaping its cost function in hindsight through a longer horizon MPC plan. We, however, are interested in tasks where cost functions are not available and cannot adopt this approach. \citet{amos2017optnet, donti2017task} also look at embedding differentiable optimization procedures (quadratic programs) within neural networks. Concurrently, a few recent efforts have been developed to embed differentiable planning procedures in computation graphs \cite{guez2018learning, pereira2018mpc, farquhar2017treeqn}. However, to our knowledge, our paper is the first to connect the use of differentiable planning procedures to learning reusable representations that generalize across complex visuomotor tasks.

The idea of planning by gradient descent has existed for decades \cite{Kelley:1960}. While such work relied on known analytic forms of environment dynamics, later work \cite{Schmidhuber90anon-line} explored jointly learning approximate models of dynamics with neural networks. \citet{henaff2017model} adopt gradient-based trajectory optimization for model-based planning in discrete action spaces, but rely on representations learned from unsupervised pretraining. \citet{oh2017value} and \citet{silver2016predictron} have also explored forward predictions in a latent space that is learned by decoding the value function of a state. Our architecture is related in so far as distance to goal in the learned latent space can be viewed as a value function. However, we also differ significantly by not relying on extrinsic rewards and focusing on continuous control tasks.

Similar to our work, \citet{pathak*2018zeroshot} and \citet{nair2017combining} train goal-conditioned policies for imitation learning, by providing an image of the goal as input to the policy. However, we show in our experiments that, unlike these methods, the representation learned via our approach can be reused for planning and reward specification.

\begin{figure*}[ht]
\centering     
\subfigure[Pointmass- VOVG]{\label{fig:pointmasszeroshota}\includegraphics[width=0.235\linewidth]{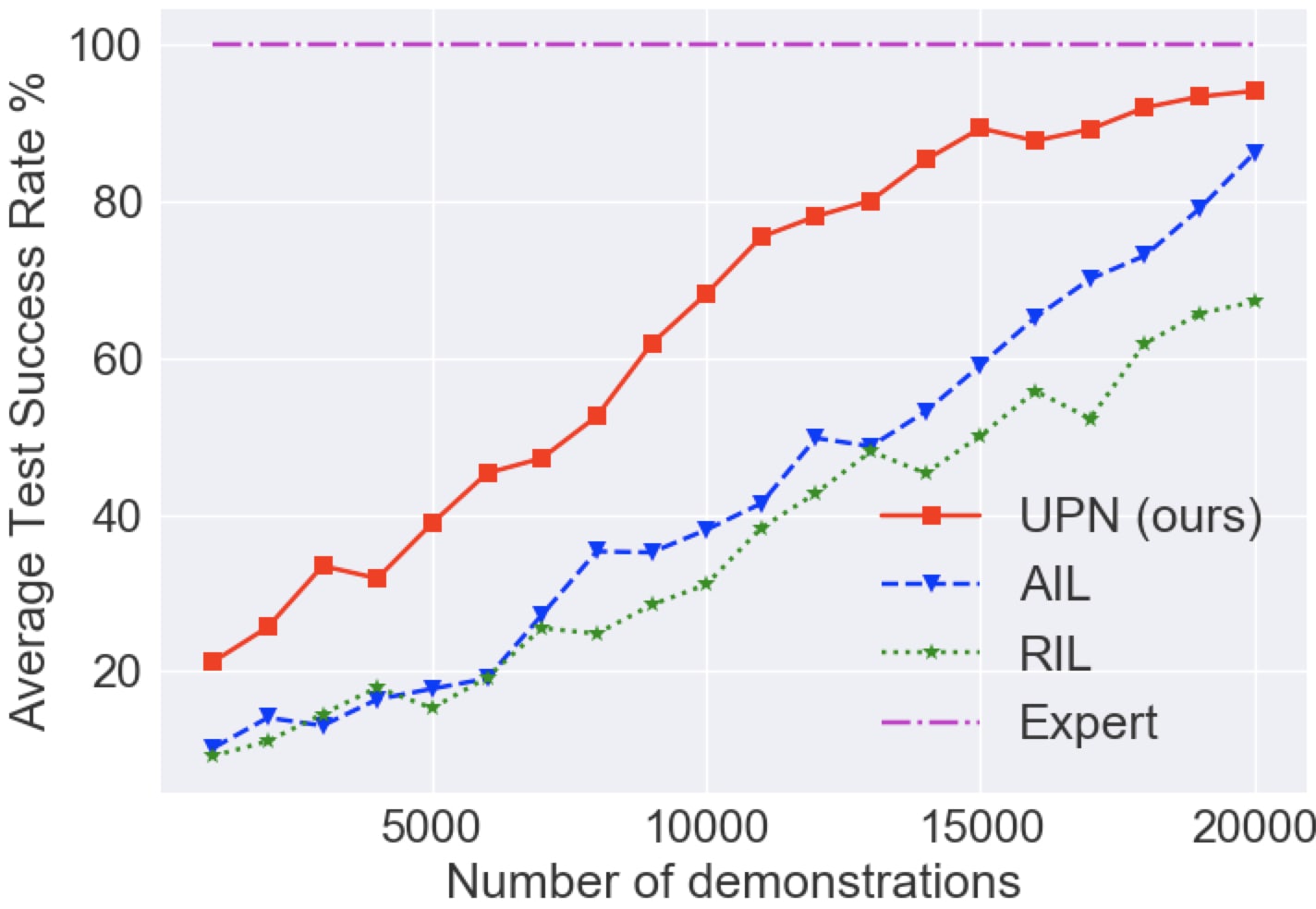}}%
\hspace{1mm}\subfigure[Reacher - VOVG]{\label{fig:reacherzeroshota}\includegraphics[width=0.235\linewidth]{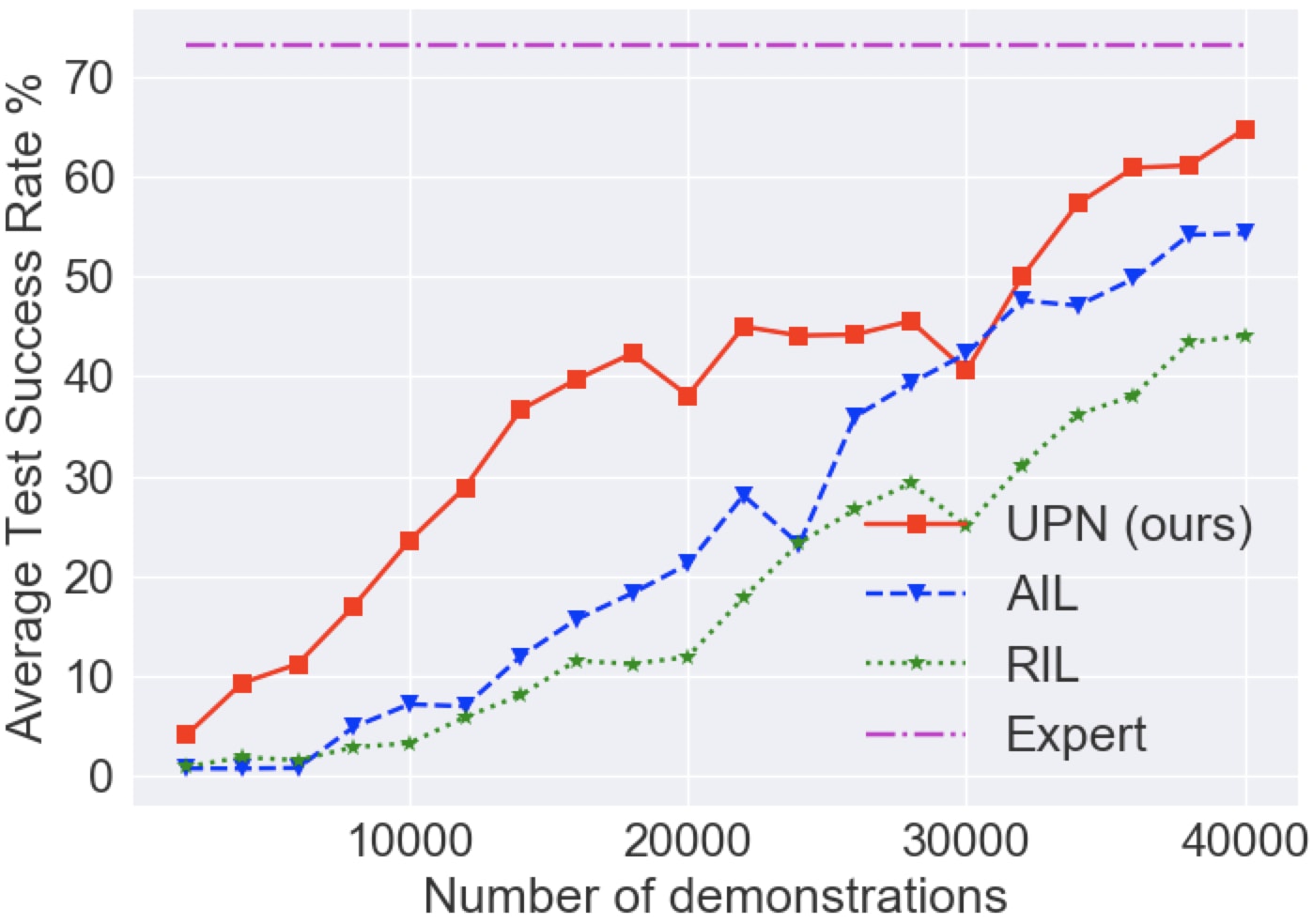}}
\hspace{1mm}\subfigure[Pointmass- FOVG]{\label{fig:pointmasszeroshotb}\includegraphics[width=0.235\linewidth]{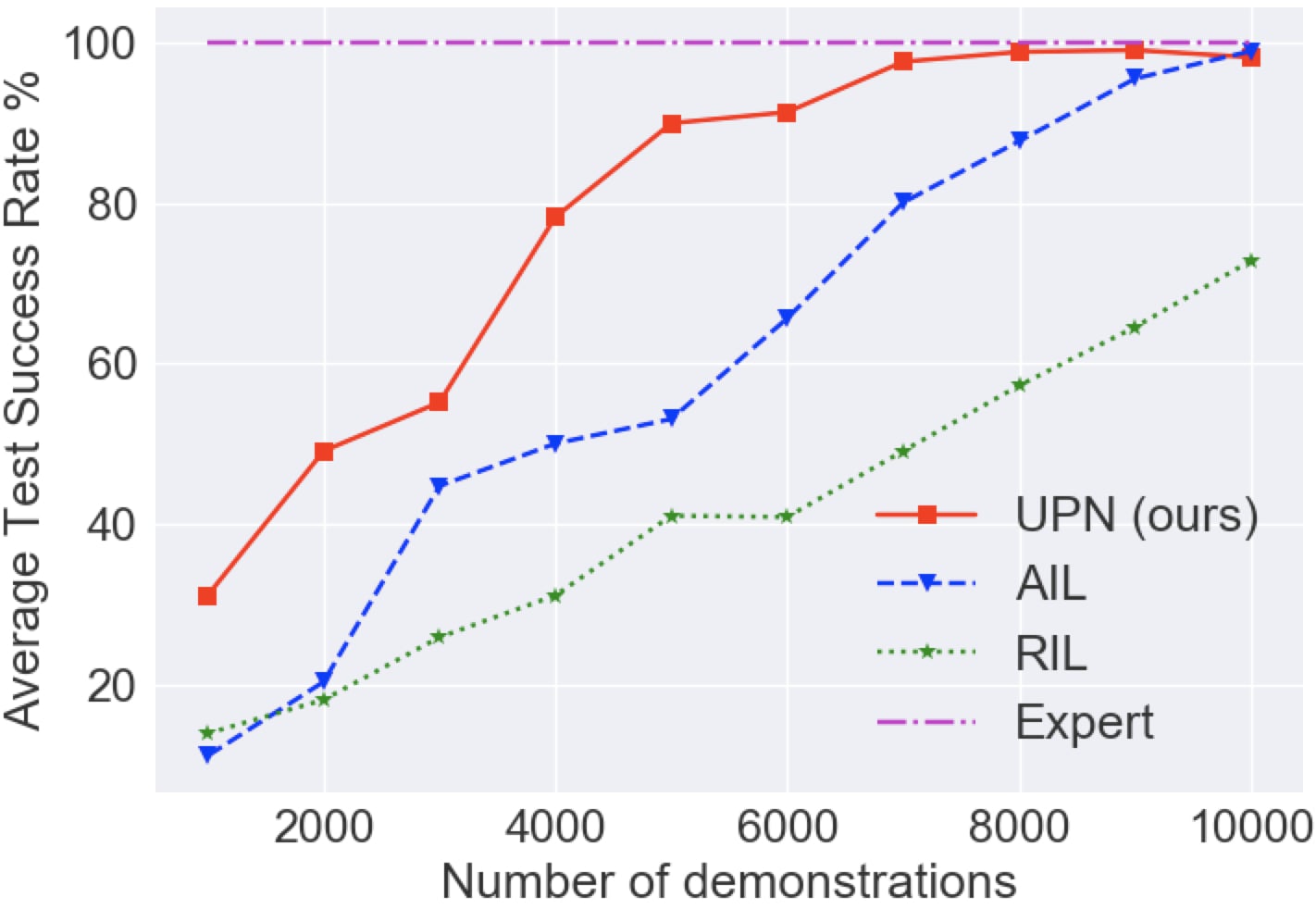}}
\hspace{1mm} \subfigure[Reacher - FOVG]{\label{fig:reacherzeroshotb}\includegraphics[width=0.235\linewidth]{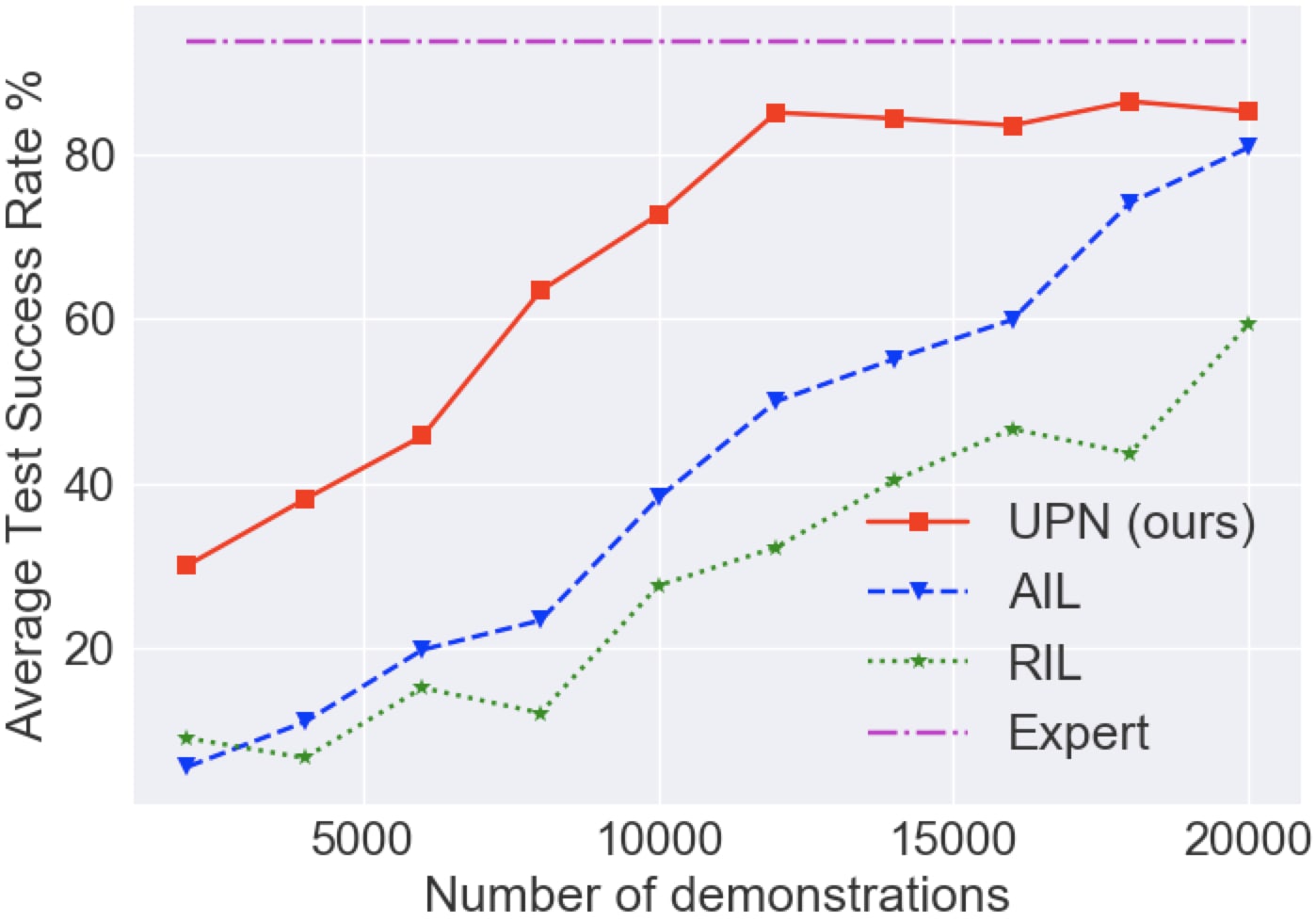}}

\caption{{\textbf{Notation:}} VOVG - Varying Obstacles and Varying Goals, FOVG - Fixed Obstacles and Varying Goals; Success on test tasks as a function of the dataset size. Our approach (UPN) outperforms the RIL and AIL consistently across the four generalization conditioned considered and is  more sample efficient. As expected, the AIL improves with more data to eventually almost match the UPN. This illustrates the tradeoff between inductive bias and expressive architectures when given sufficient data.}
\vspace{-2mm}
\label{fig:zeroshotresults}
\end{figure*}

\section{Experiments}

We designed experiments to answer the following questions: (1) does embedding a gradient descent planner help learn a policy that can map from pixels to torque control when provided current and goal observations at test-time ? (2) how does our method compare to reactive and autoregressive behavior cloning agents as the amount of training data varies? (3) what are the properties of the representation learned by UPN? (4) how can the learned representations from UPN be leveraged for transfer to new and more complex tasks, compared to representations from standard imitation methods and unsupervised methods (e.g. VAE)? 

\textbf{Methods for comparison:} We consider two alternative imitation learning approaches for comparison:
(1) a reactive imitation learner (RIL), composed of a convolutional feedforward policy that takes as input the current and goal observation;
(2) an auto-regressive imitation learner (AIL), composed of a recurrent decoder initially conditioned on convolutionally encoded representations of the current and goal observation, trained to output a sequence of intermediate actions. Both (1) and (2) are methods adopted from \citet{pathak*2018zeroshot}. These comparisons are important for studying the effects  of the inductive bias of gradient descent planning that is embedded within UPN. More specifically, comparing to (1) allows us to understand the need for such an inductive bias, while comparing to (2) is necessary to understand whether the benefits are not purely due to recurrent computations. All methods are trained on the same synthetically-generated expert demonstration datasets. We refer the reader to the supplementary for details on the architectures and dataset generation.


\subsection{UPNs Learn Effective Imitation Policies} \label{upnzeroshot}

Here, we study the suitability of the UPN for learning visual imitation policies that generalize to new goal-directed tasks. We focus on two tasks: (1) navigating a 2D point robot around obstacles to desired goal locations amidst distractors (Figures \ref{fig:pointmassa} and \ref{fig:pointmassb}), wherein the color of the goal is randomized; (2) a harder task of controlling a 3-DoF planar arm to reach goals amidst scattered distractors and obstacles, as shown in Figures \ref{fig:reachera} and \ref{fig:hardobstacle}. 

For these tasks, we consider two types of generalization: (1) generalizing to new goals for a fixed configuration of obstacles having trained on the same configuration; (2) generalizing to new goals in new obstacle configurations having trained across varying obstacle configurations. Figures \ref{fig:reachera} and \ref{fig:hardobstacle} show two different obstacle configurations for the reaching task, while the differently colored locations in Figure \ref{fig:tasksnapshots} represent varying goal locations.

We employ the action selection process described in subsection \ref{learningtoplan} with a chosen maximum episode length. Results shown in Figure \ref{fig:zeroshotresults} compare performance over a varying number of training demonstrations. 
As expected, the inductive bias of embedding trajectory optimization via gradient descent in UPN supports generalization from fewer demonstrations. With more demonstrations, however, the expressive AIL is able to almost match the performance of the UPN. 
This is consistent with the conclusions of \citet{tamar2016value}, who observed that the benefit of the value iteration inductive bias shrinks in regimes in which demonstrations are plentiful. Note that generalization across obstacle configurations in the reacher case (Figure \ref{fig:reachera}) is a hard task; expert performance is only 73.12\%. We encourage the reader to refer to the supplementary for further details about the experiment.

\begin{figure}[h]
\centering

\subfigure[]{\label{fig:planstepsa}
\includegraphics[width=0.235\textwidth]{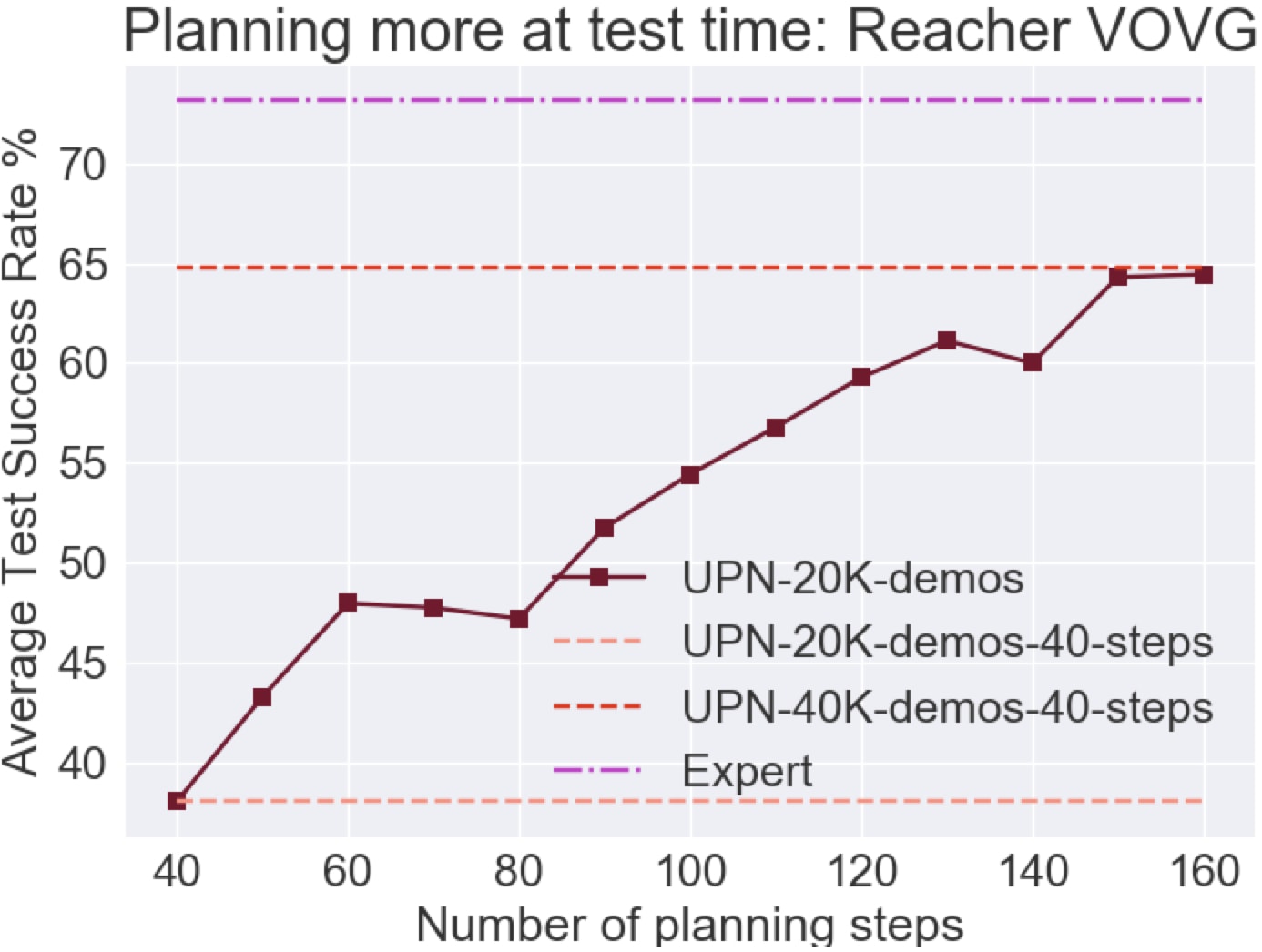}} \hspace{-2mm}\
\subfigure[]{\label{fig:planstepsb}
\includegraphics[width=0.235\textwidth]{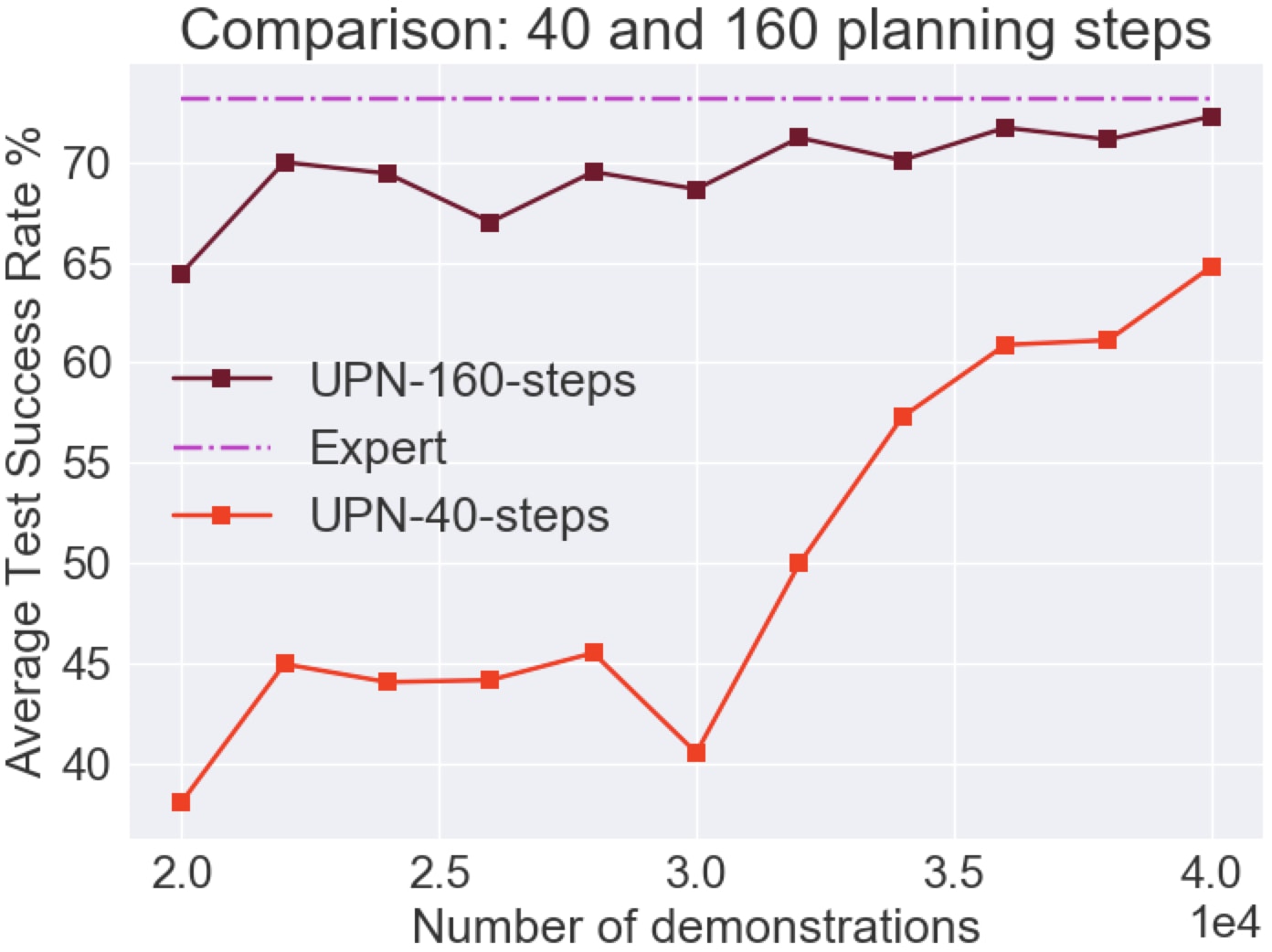}}
\vspace{-5mm}
\caption[width=0.2\textwidth]{\textbf{(a)} The effect of additional planning steps at test-time. UPN learns an effective gradient descent planner whose convergence improves with more planning steps at test-time. 
\textbf{(b)} A comparison of the success rate of UPN between 40 and 160 planning steps at test time with varying number of demonstrations on Reacher VOVG. Using 160 planning steps is consistently better than using 40 steps (though the relative benefit shrinks with more demonstrations) and allows the UPN to match the expert level. 
} 

\label{fig:plansteps}
\end{figure}

\subsection{Analysis of the Gradient Descent Planner} \label{upnplansteps}

The UPN can be viewed in the context of meta-learning as learning a planning algorithm and its underlying representations. 
We take inspiration from \citet{finn2017meta}, who studied a gradient-based model-agnostic meta-learning algorithm and showed that a classifier trained for few-shot image classification improves in accuracy at test-time with additional gradient updates.
In our case, the inner loop is the GDP, which may not necessarily converge due to the fixed number of planning updates. 
Hence, it is worth studying whether additional test-time GDP updates yield more accurate plans and therefore better success rates.

\textbf{Planning more helps:} 
Figure \ref{fig:planstepsa} shows that with more planning steps at test-time, a UPN trained with fewer demonstrations (20000) can improve on task success rate beginning from 38.1\% with 40 planning steps to 64.44\% with 160 planning steps. As a reference, the average test success rate of the expert on these tasks is 73.12\% while the best UPN model with 40 planning steps (trained on twice the number of demos (40000)) achieves 64.78\%. Thus, with more planning steps, we see that UPN can improve to match the performance of a UPN with fewer planning steps but trained on twice the number of demonstrations.
We also find that 160 steps is consistently better
than using 40 steps (though the relative benefit shrinks with more
demonstrations) and that the UPN is able to match expert performance (Figure \ref{fig:planstepsb}). This finding suggests that the learned planning objective is well defined, and can likely be reused for related control problems, as we explore in Sections~\ref{harderscenarios} and~\ref{morphtransfer}.

\subsection{Latent Space Visualization}
\begin{figure}[h]
\centering
\includegraphics[width=0.5\linewidth]{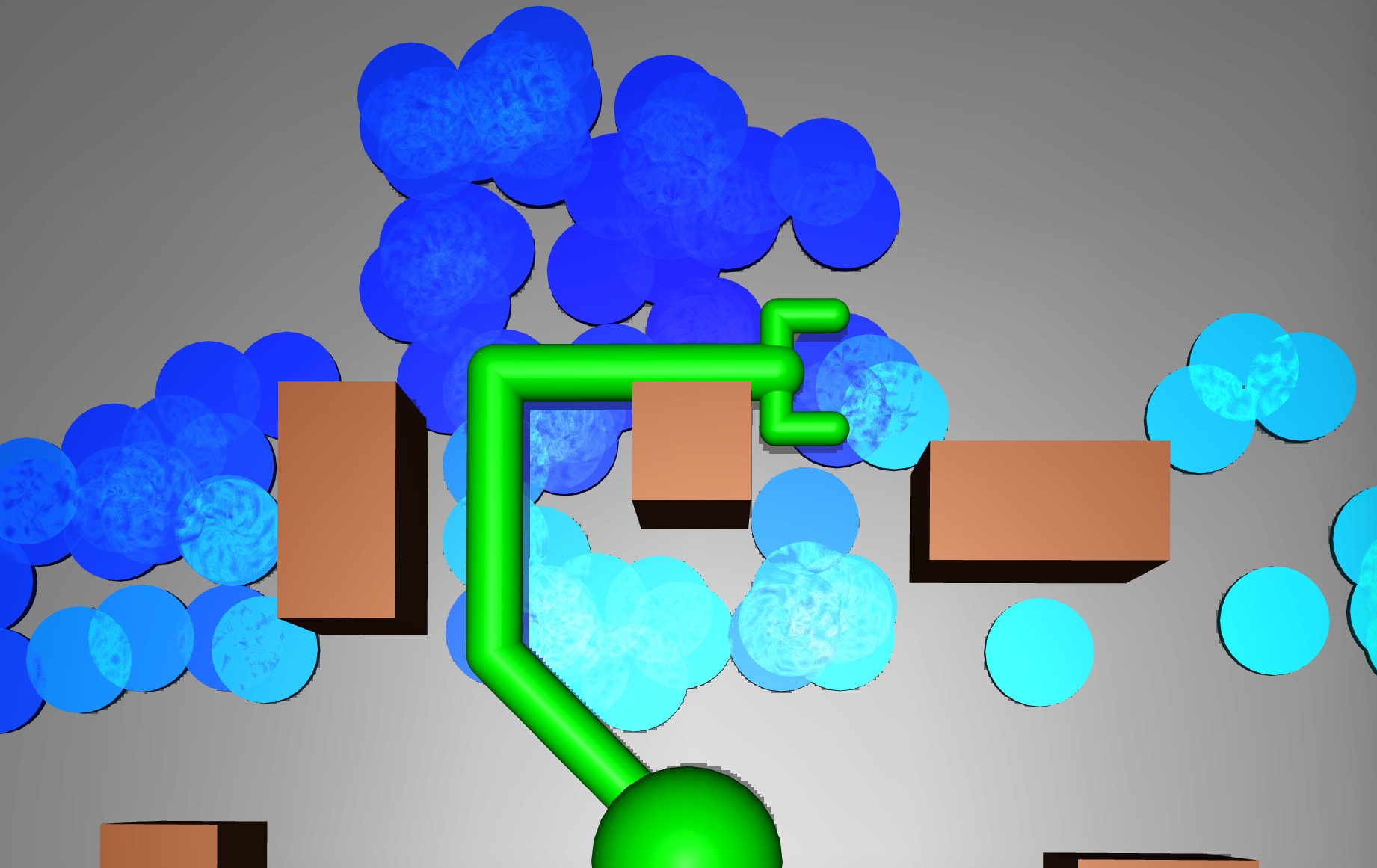} \
\vspace{-3mm}
\caption[width=0.2\textwidth]{ Visualization of the learned metric in the UPN latent space on the reacher with obstacles task. Lighter color $\rightarrow$  larger latent distance. The learned distance metric is obstacle-aware and supports obstacle avoidance. }
\vspace{-3mm}
\label{fig:featviz}
\end{figure}

We offer a qualitative analysis for studying the acquired latent space for an instance of the reacher with obstacles task.
Given the selected initial pose, we compute the distance in the learned $f_\phi$ space for 150 random final poses and illustrate these distances qualitatively on the environment arena by color mapping each end-effector position accordingly. 
The result is shown in Figure \ref{fig:featviz}; lighter blue corresponds to larger distances in the feature space.  We see that the learned distance metric is \emph{obstacle-aware} and task-specific: regions below the initial position in Figure \ref{fig:featviz} are less desirable even though they are near, while farther regions above are comparatively favorable.

\begin{table}[ht]
\footnotesize
\caption{Average Success Rate \% in solving the task described in Figure \ref{fig:hardobstacle} for fixed and varying goals}
\vspace{-1mm}
\label{sample-table}
\vskip 0.15in
\begin{center}
\begin{small}
\begin{sc}
\begin{tabular}{lcccr}
\toprule
Feature Space & Fixed & Varying \\
\midrule
RIL-RL    & 0\% & 0.01\%\\
AIL-RL & 0\% & 4.72\%\\
VAE-RL    & 20.23\% & 24.67\%\\
UPN-160 Imitation & 45.82\% & 47.99\% \\
Expert & 46.77\% & 51.1 \% \\
{\textbf{UPN-RL}}    &     {\textbf{69.84\%}} & {\textbf{71.12\%}}     \\
\bottomrule
\end{tabular}
\end{sc}
\end{small}
\end{center}
\vskip -0.1in
\end{table}

\begin{figure*}[ht]
\vspace{-2mm}
\centering

\subfigure[Transfer to more complex topology (Reacher)]{\label{fig:topologytransfer}
\includegraphics[width=0.45\linewidth]{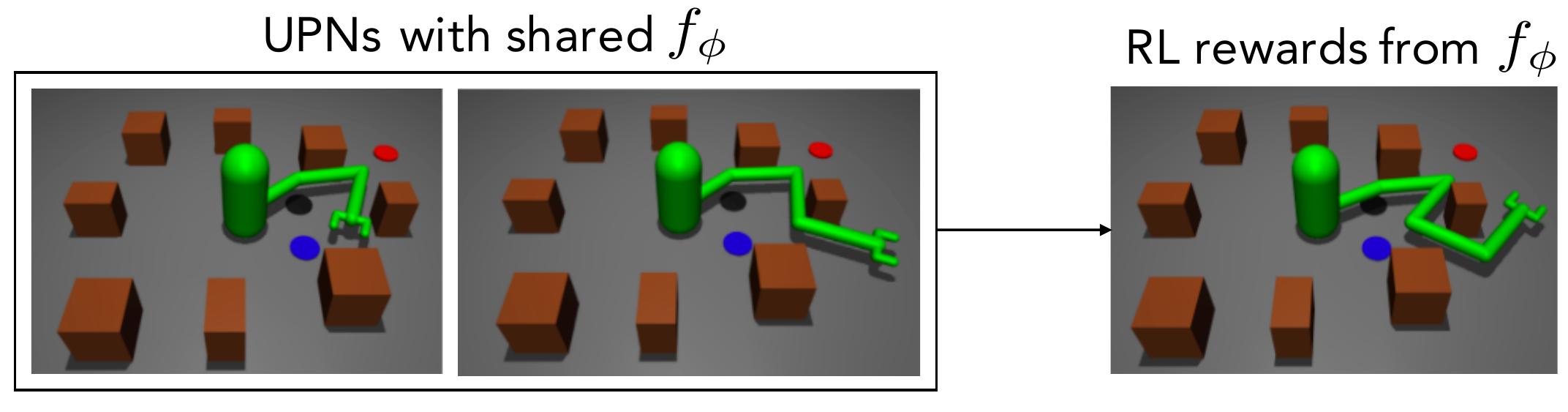}} \
\subfigure[Transfer to new morphology (Point to Ant)]{\label{fig:morphologytransfer}
\includegraphics[width=0.40\linewidth]{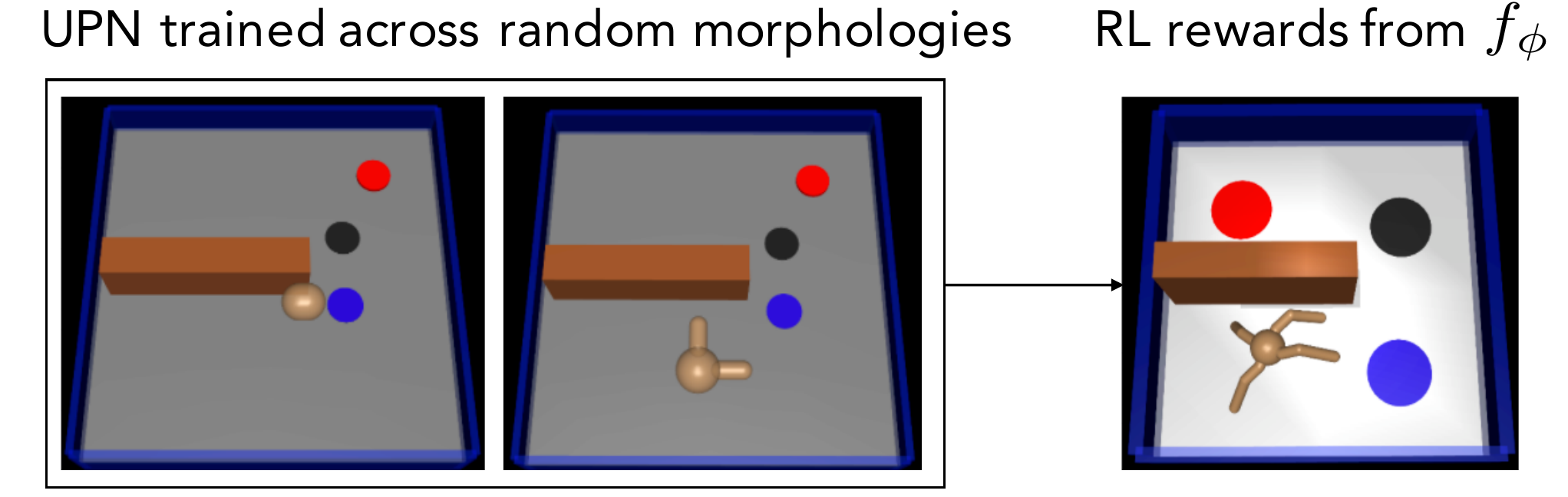}}

\vspace{-4mm}
\caption{Transfer between robots as described in subsection 4.5.}
\vspace{-3mm}
\label{fig:transfer}
\end{figure*}{}

\begin{figure*}[ht]
\centering
\subfigure[Point robot to Ant transfer]{\includegraphics[width=0.25\textwidth]{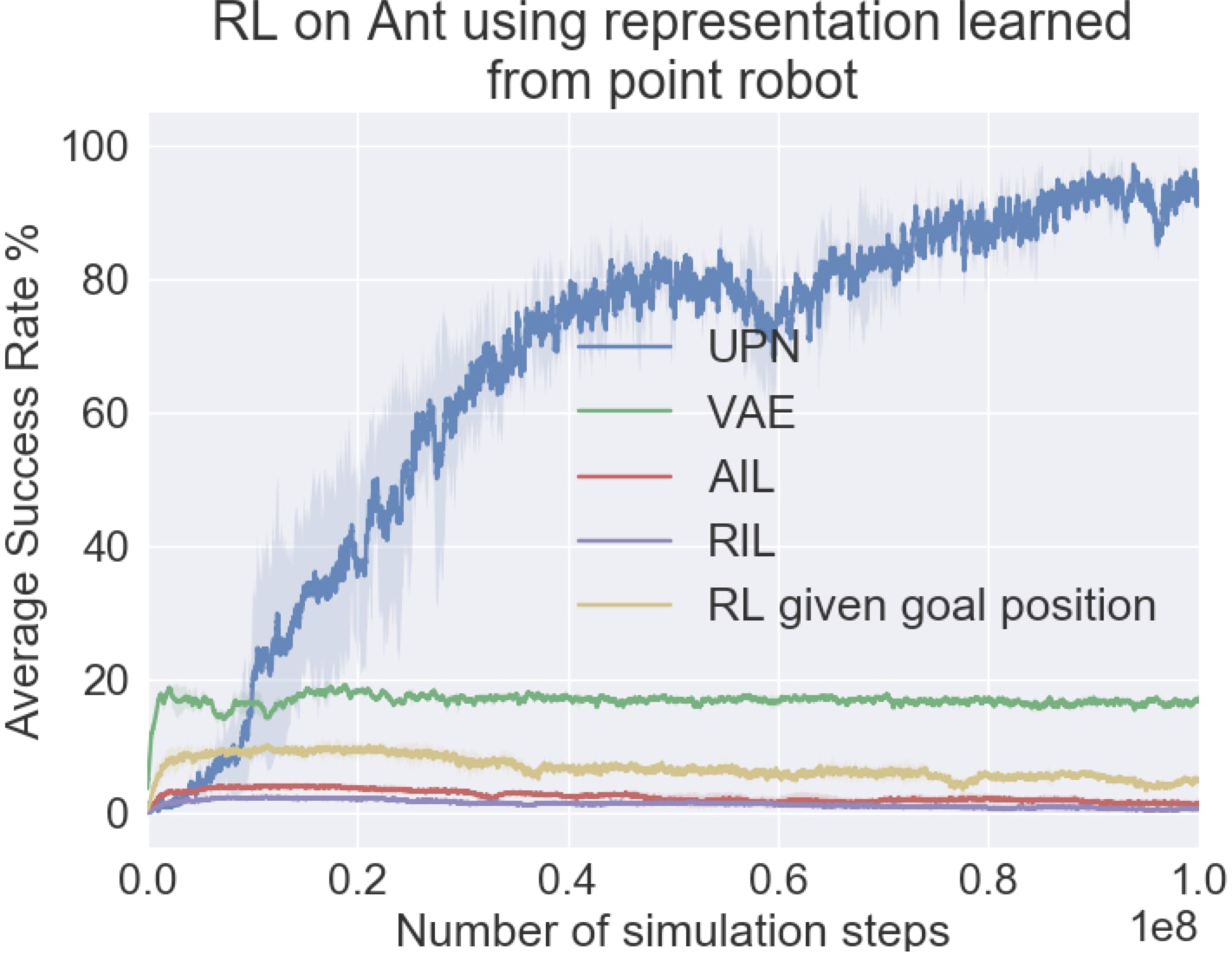}\label{fig:antresult}
\label{fig:antrl1}} \hspace{-1mm}    \
\subfigure[Reacher transfer]{\includegraphics[width=0.25\textwidth]{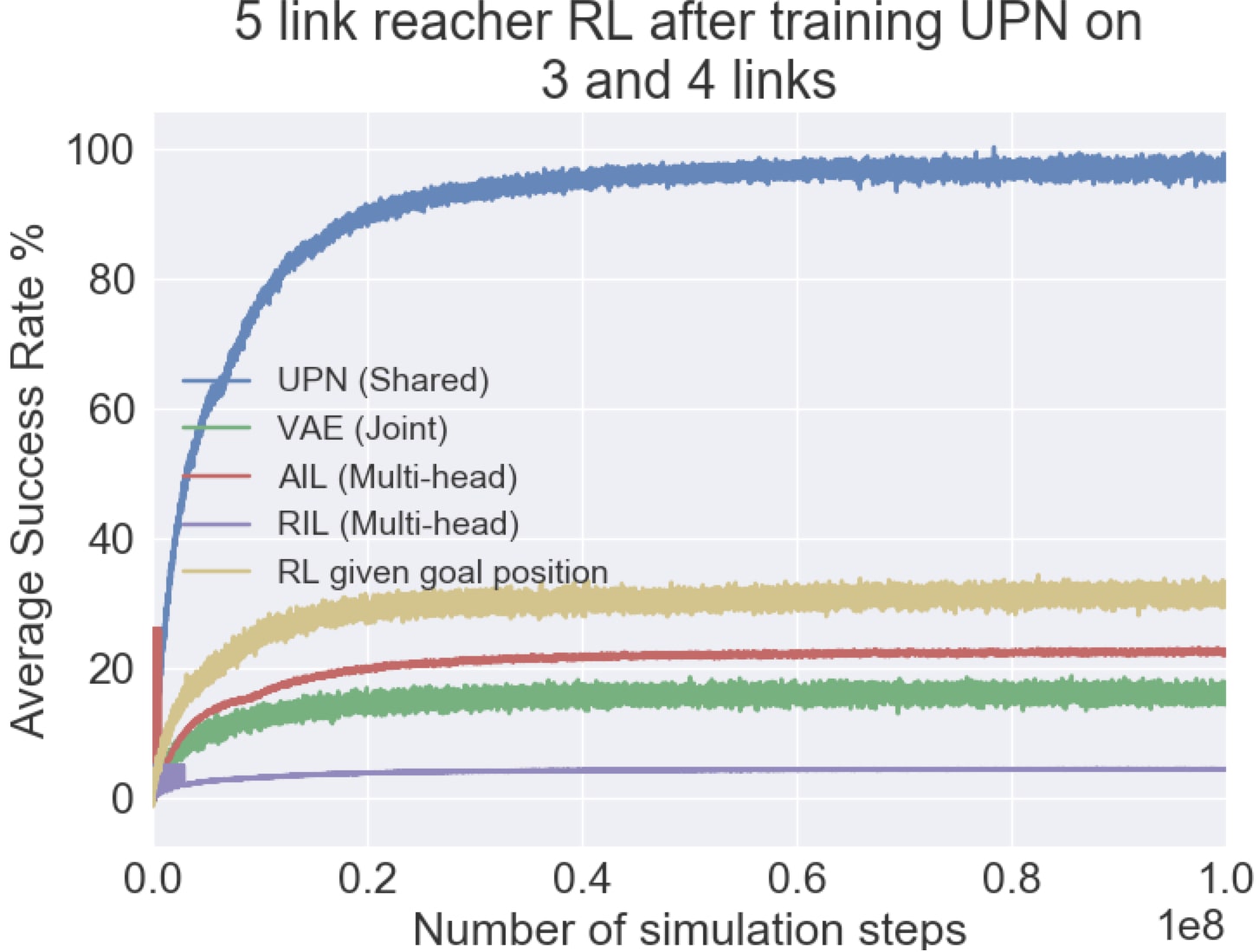}\label{fig:reachertransferresult}
\label{fig:antrl2}} \hspace{-1mm}   \ 
\subfigure[Pushing from poking transfer]{    \includegraphics[width=0.25\textwidth]{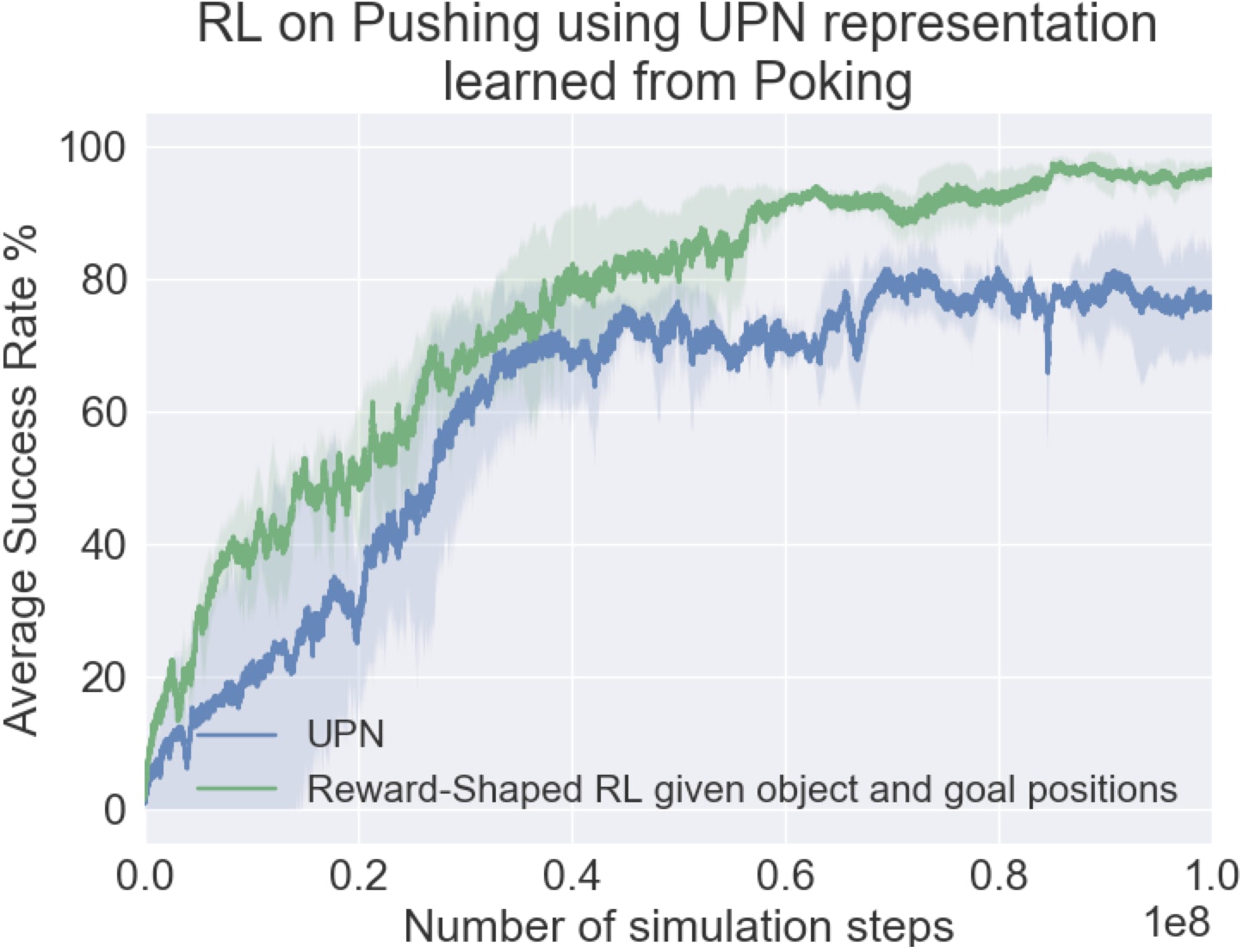}
\label{fig:pushresult}}\hspace{-1mm}  \ 
\subfigure[7-DoF Pushing task]{\includegraphics[width=0.20\textwidth]{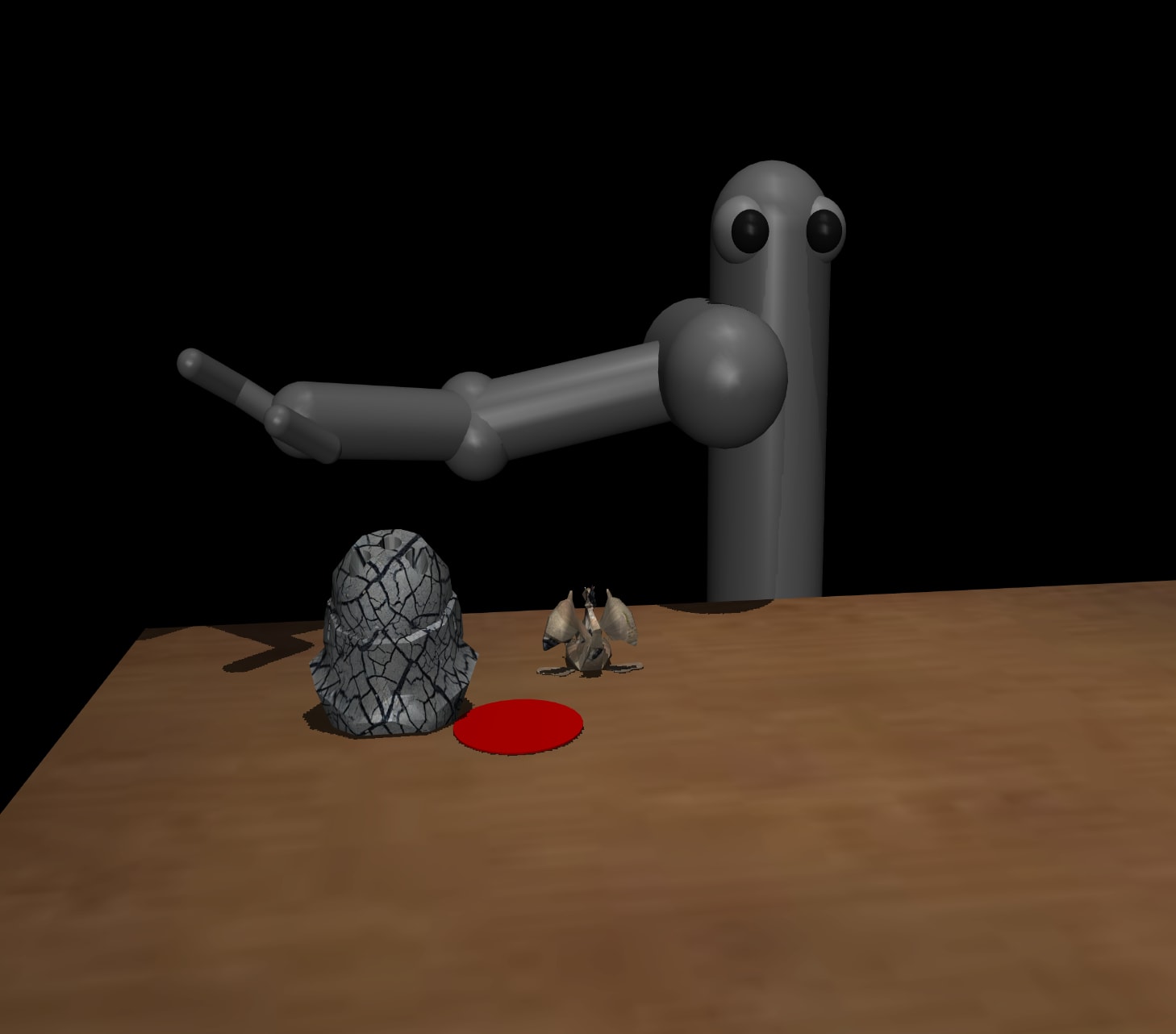} \label{fig:push}}
\vspace{-4mm}
\caption{(a-c) RL with rewards from the UPN representation is significantly more successful compared to other feature spaces (VAE, AIL, RIL, shaped rewards), suggesting that UPNs learn transferrable, generalizable latent spaces.}
\vspace{-4mm}
\label{fig:transferplots}
\end{figure*}{}

\subsection{Transfer to Harder Scenarios} \label{harderscenarios}

We have seen in subsections \ref{upnzeroshot} and \ref{upnplansteps} that UPNs can learn effective imitation policies that can perform close to the expert level on visuomotor planning tasks. 
In principle, deriving reward functions from a trained UPN as explained in subsection \ref{rlexplanation} should allow us to extend beyond the capabilities of the expert on harder scenarios where the expert fails. We study this idea in the reaching scenario with obstacle configuration as presented in Figure \ref{fig:hardobstacle}. The difference between fixed and varying goals is that for varying goals, we feed in a feature vector of the goal image as an additional input to the RL policy. We use PPO \cite{schulman2017proximal} for model-free policy optimization of the rewards derived from the feature space(s). Though the subsection \ref{rlexplanation} explains the reinforcement learning procedure in the context of using $f_\phi$ from a UPN, one could use a trained encoder $f_\phi$ from other methods such as our supervised learning comparisons RIL, AIL. In addition to RIL and AIL, a feature space we compare to is an encoder obtained from training a variational auto-encoder (VAE) \cite{kingma2013auto} on the images of the demonstrations.  This comparison is necessary to judge how useful the feature space of a UPN is for downstream reinforcement learning when compared to pixel reconstruction methods such as VAEs. In Table~\ref{sample-table}, we see that reinforcement learning on the feature space of RIL and AIL clearly fail, while RL on the UPN feature space is significantly better compared to that of a VAE. We also see that UPN-RL is able to outperform the expert and the imitating UPN-160.

\subsection{Transfer Across Robots} \label{morphtransfer}

Having seen the success of reinforcement learning using rewards derived from UPN representations in subsection \ref{harderscenarios}, we pose a harder problem in this subsection: Can we leverage UPN representations trained on some source task(s) to provide rewards for target task(s) with significantly different dynamics and action spaces? We propose to do this by training and testing with different robots (morphological variations) on the same desired functionality (reaching / locomotion, around obstacles). This study will highlight the extrapolative nature of UPN representations. The idea of trajectory optimization with a learned metric is a fundamental prior that can hold across a large class of visuomotor control problems. Having trained UPN to learn such a prior, it is natural to expect the underlying representation to be amenable to providing suitable metric based rewards for similar but unseen tasks. We craft two challenging experimental scenarios to verify this hypothesis.

\textbf{Reacher with new morphology:} Having trained a UPN with a shared $f_\phi$ and different $g_\theta$ for a 3-link and 4-link reacher (on the obstacles task), can we leverage the learned $f_\phi$ to specify rewards for reinforcement-learning a 5-link reacher to reach different goals around the same obstacles? Figure \ref{fig:topologytransfer} visually depicts this experiment. Such a transfer scenario hasn't been studied in the past for visuomotor control. The dynamics of a 5-link reacher are more complex (compared to 3 and 4 link reachers), thereby posing a harder control problem to solve at test time. However, a good path-planning reward function learned from 3 and 4-link reachers is likely to help for a 5-link reacher due to morphological similarities. We train the UPN on both the 3 and 4 link reachers to avoid overfitting the learned metric to a specific dynamical system. As comparison methods, we train RIL and AIL (with a multi-task (head) architecture), and a VAE (jointly on images from both the tasks).

\textbf{Point to Ant:} Higher-level navigation to goals amidst obstacles should be common across different robots, from a 2D point robot controlled through simple forces to a robot as complex as an 8-joint quadruped ant. While the lower level actuation varies across different robots, the visual spatial planning should ideally be transferrable . We empirically confirm this via an experiment illustrated in Figure \ref{fig:morphologytransfer}. Here, we learn representations with a UPN on demonstrations collected from a 2D point robot trained to traverse obstacles to reach varying goals. We randomize the robot's morphological appearance across demonstrations (Figure \ref{fig:morphologytransfer}), inspired by \citet{sadeghi2016cad2rl,tobin2017domain}. This allows the UPN to learn an encoder $f_\phi$ that is robust to the creature appearance. We then design an experiment to use this $f_\phi$ for a harder problem. First, we train a UPN with a simple 2D point robot; we then replace the point robot with a 3D-torque-controlled ant, which requires more delicate handling of the surface contacts for maneuvering the quadruped and avoiding obstacles. Once again, to our knowledge, such morphological transfer has not been demonstrated in prior work on visuomotor control.

In Figure \ref{fig:transferplots} we see that for both the transfer scenarios, RL with rewards from the UPN representation is significantly more successful compared to other feature spaces (VAE, AIL, and RIL). In addition to other feature spaces, we also compare the UPN-RL setup to a naïve RL agent optimizing a spatial distance to goal in the co-ordinate space as the reward; this procedure assumes that the spatial position of the goal is known, unlike UPN, RIL, AIL, and VAE where the feature vector of the goal image is provided as input. We note that this method also performs poorly, which is expected because, unlike distances in the feature space of UPN, the spatial distance in the co-ordinate space is not obstacle-aware. Note that UPN-RL relies only on the UPN representation to provide the RL agent with knowledge of the task, thus inferring the goal from the obstacle-informative latent space. These results show that optimizing for the rewards derived from UPN correlates with task success, supporting our claim that UPNs learn {\it generalizable} and {\it transferrable} latent spaces. We show further extrapolation (6-link and 7-link reachers) in our video results. To our knowledge, there has been no prior exposition of torque controlled goal conditioned ant navigation for varying goals around obstacle(s) even when spatial positions of the goal and ant torso are known. UPN is therefore an effective way of uncovering useful metric priors that can serve as perceptual reward functions for complex tasks for which reward functions are typically hard to engineer.

\subsection{3D 7-DoF Control from an Non-orthographic View}

So far, we have demonstrated results with UPN on tasks where the view point is orthographic, which may make it easier for the agent to map from pixels to relative positions. We next seek to answer the question: Can UPN still work for scenarios where the camera view of the task is non-orthographic? This is a common scenario for real robot tasks or more complex manipulation tasks in simulation \cite{finn2017one}. To answer this, we consider the task of controlling a 3D 7-DoF arm from \textit{non-orthographic} viewpoints, which presents a harder perception problem (shown in Figure \ref{fig:push}). This task is adapted from \citet{finn2017one} where there is a distractor object in addition to a target object that needs to be displaced to a goal location. However, here, we do not focus on generalization to new objects unlike \citet{finn2017one}. Instead, we look at skill generalization. We collect a dataset of random pokes (see \citet{agrawal2016learning} for a detailed description of poking in robotics and our video highlights for visual illustration of poking trajectories). Having trained a UPN representation on poking trajectories, we study whether rewards derived from it can guide learning of more complex and composite skills, such as pushing (which involves appropriately reaching for the target object and guiding it to the goal). Further, we also analyze whether the UPN representation based rewards can replace hand-engineered reward shaping for such a task. 

Having established the clear success of UPN over RIL, AIL, and VAE for the RL experiments in subsections \ref{harderscenarios} and \ref{morphtransfer}, we perform a different comparison here. In addition to UPN-RL, we train the agent on the pushing task assuming the object and goal positions are known using RL without image based inputs and a well shaped reward that is described \href{https://github.com/openai/gym/blob/master/gym/envs/mujoco/pusher.py#L19}{here}. To our surprise, we find that the transfer from poking to pushing using the UPN-RL setup works efficiently. Our method, which relies only on UPN representation rewards and the current image, approaches the performance of the shaped reward function in terms of task success  (Figure \ref{fig:pushresult}). This result suggests that the UPN latent space captures the proximity of the end-effector to the object as a pre-requisite to moving the object to desired locations and hence supports acquisition of even more complex behavior via reinforcement learning. Thus, UPN can serve as a means to acquiring a structured metric in the latent space from which reward functions for complex manipulation tasks such as pushing can emerge naturally.

\begin{figure}[ht]
\vspace{-2mm}
\centering

\subfigure[Humanoid Task]{\label{fig:humanoid}
\includegraphics[width=0.45\linewidth]{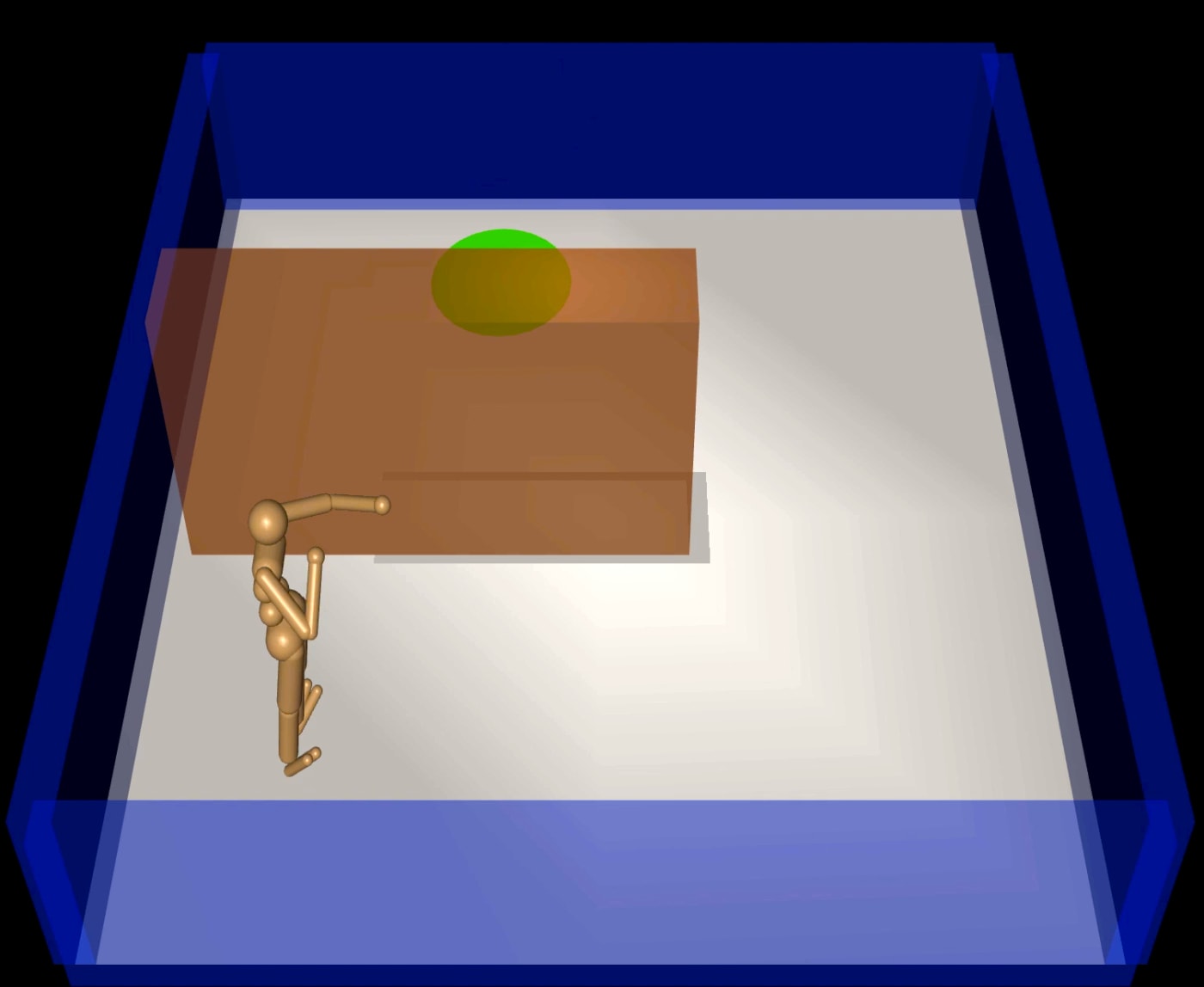}} \
\subfigure[Ant: Long Horizon Task]{\label{fig:antlong}
\includegraphics[width=0.45\linewidth]{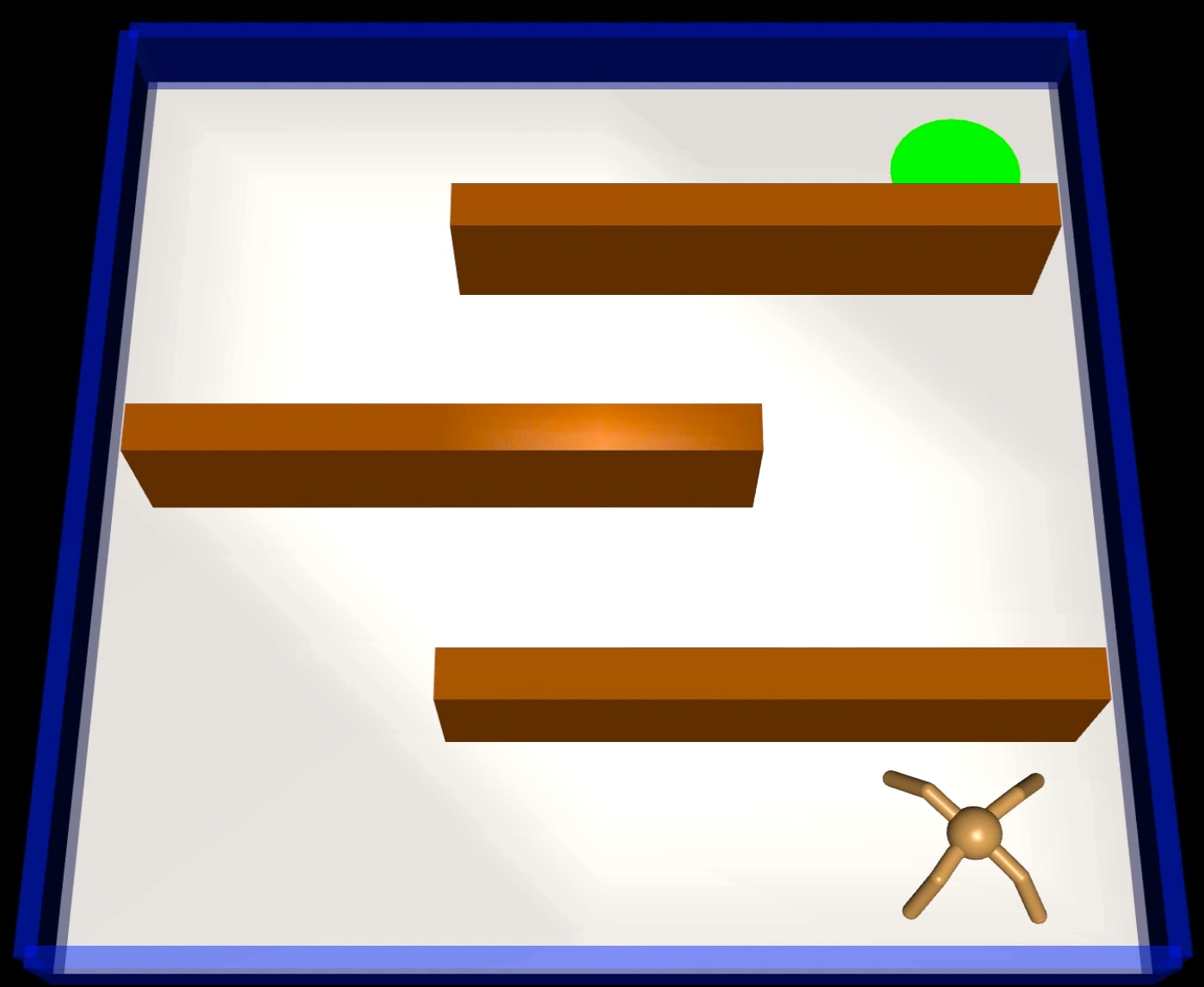}}

\vspace{-4mm}
\caption{Using UPN representations trained on a simple 2D point robot, we control complex robots such as a full humanoid and an ant to locomote around obstacles and reach the green goal. While the humanoid task is challenging because of the complex actuation, the ant task requires optimizing a policy over 8000 time steps, providing evidence that the rewards from UPN representations can aid in effective credit assignment over long horizons.} 
\vspace{-3mm}
\label{fig:newresults}
\end{figure}{}

\subsection{Transfer from Point Robot to a Humanoid: Pushing the limits of generalization} \label{humanoidsec}

In this subsection, we consider the following question: Can a reward function derived from a very simple creature such a 2D point robot be used to control a much more complex robot such as a humanoid, for similar behaviors such as locomotion around obstacles (ref Figure \ref{fig:humanoid})? This is a harder problem than transferring reward functions from point robot to ant because a humanoid is a much more complex robot to control. To deal with the visual differences between a humanoid and a point robot, we assume that we have access to the 2D co-ordinate position of the center of mass of the humanoid. We re-render the humanoid as a 2D point robot whose center of mass is same as that of the humanoid, and pass this image through a UPN trained on a point robot to provide the reward function for the humanoid. Note that this assumption of knowing the global location is very minimal, since the reward function is still non-trivial, and in this case, a learned perceptual metric. 

In the case of the quadruped, it wasn't necessary to shape the rewards for balancing as long as we could terminate episodes whenever the creature falls down. However, for a humanoid, an explicit reward for staying on feet is necessary. We saw that naive termination of episodes on falling down resulted in the humanoid moving close to the side walls to balance itself and stay on feet throughout the duration of the task, rather than optimizing for the path planning reward derived from the UPN representations. To get around this issue, we adopted the strategy used in \citet{bansal2017emergent}, whereby a decaying curriculum for staying on feet is used, and combined with the metric reward for the shortest path locomotion. Check out the video of the behavior learned by the humanoid on the project webpage: \url{https://sites.google.com/view/upn-public}.

\subsection{Using UPN rewards for long horizon tasks} \label{antlongsec}

Finally, we answer: To what extent can the learned metrics from UPN guide a new reinforcement learning agent? One way to push the limits is to consider long horizon tasks. Typically, continuous control tasks studied in reinforcement learning are restricted to a maximum horizon of less than or equal to 1000 simulation steps. Taking inspiration from \citet{frans2018meta}, we study a goal conditioned navigation task wherein a quadruped has to locomote around obstacles to reach a goal that is far away (time horizon of 8000 simulation steps, refer to Figure \ref{fig:antlong}). In the case of \citet{frans2018meta}, the reward was extrinsic and sparse, and thus a hierarchical policy was required for efficient credit assignment and exploration. In our case, we show that the shaped reward from UPN on a point robot can overcome the problem of sparse rewards and effective credit assignment. This experiment indicates that UPN is able to learn efficient distance metrics in an abstract space from very simple short horizon tasks such as controlling a 2D point robot, which can be powerful enough to guide the reinforcement learning of complex policies such as controlling an ant to move around mazes over much longer horizons. Check out the video of the learned behavior on the project webpage: \url{https://sites.google.com/view/upn-public}.

\vspace{-2mm}

\section{Discussion}

We posed the problem of learning representations for performing generalizable visuomotor control. We focused on one property such a space should satisfy: providing distance metrics for reinforcement learning on tasks specified via goal images without  extrinsic rewards. To this end, we introduced universal planning networks, a goal-directed policy architecture with an embedded differentiable planner, that can be trained end-to-end. Our extensive experiments demonstrated that (1) UPNs learn effective visual goal-directed policies efficiently; (2) UPN latent representations can be leveraged to transfer task-related semantics to more complex agents and more challenging tasks through goal-conditioned reward functions; and (3) the learned planner improves with more updates at test-time, providing encouraging evidence of meta-learning for planning. Our transfer learning successes demonstrate that we have learned generic representations that have notions of agency and planning. 
Future work should investigate different ways to train UPN representations, such as through reinforcement learning or self-supervision, borrowing ideas from \citet{andrychowicz2017hindsight}, \citet{sukhbaatar2017intrinsic}, \citet{weber2017imagination} and \citet{ pathakICMl17curiosity}. Another important future direction is to study representations wherein the metrics are structured as value functions instead of rewards as a consequence of which long horizon policy optimization could be more effective and sample efficient. Further, the results from subsections \ref{humanoidsec} and \ref{antlongsec} suggest that UPN like architectures might be practically applicable for learning complex real robotic behaviors by leveraging simulated behaviors of much simpler robots performing the same tasks.

\section{Acknowledgements}

AS thanks Aviv Tamar for insightful feedback on planning that motivated the experiments; John Schulman, Ashvin Nair and Deepak Pathak for helpful discussions and feedback; Rocky Duan for technical help; and Jonathan Ho, Jason Peng, and Kelvin Xu for reviewing previous versions of the paper. AJ thanks Alyosha Efros for support and feedback. This research was supported by an ONR PECASE N000141612723AS grant, NVIDIA and Amazon Web Services. AS and AJ were supported by a Berkeley EECS Fellowship and a Berkeley AI Research Fellowship; CF was supported by an NSF GRFP award.

\bibliography{example_paper}

\begin{thebibliography}{63}
\providecommand{\natexlab}[1]{#1}
\providecommand{\url}[1]{\texttt{#1}}
\expandafter\ifx\csname urlstyle\endcsname\relax
  \providecommand{\doi}[1]{doi: #1}\else
  \providecommand{\doi}{doi: \begingroup \urlstyle{rm}\Url}\fi

\bibitem[Abadi et~al.(2016)Abadi, Agarwal, Barham, Brevdo, Chen, Citro,
  Corrado, Davis, Dean, Devin, et~al.]{abadi2016tensorflow}
Abadi, Mart{\'\i}n, Agarwal, Ashish, Barham, Paul, Brevdo, Eugene, Chen,
  Zhifeng, Citro, Craig, Corrado, Greg~S, Davis, Andy, Dean, Jeffrey, Devin,
  Matthieu, et~al.
\newblock Tensorflow: Large-scale machine learning on heterogeneous distributed
  systems.
\newblock \emph{arXiv preprint arXiv:1603.04467}, 2016.

\bibitem[Abbeel \& Ng(2004)Abbeel and Ng]{abbeel2004apprenticeship}
Abbeel, Pieter and Ng, Andrew~Y.
\newblock Apprenticeship learning via inverse reinforcement learning.
\newblock In \emph{Proceedings of the twenty-first international conference on
  Machine learning}, pp.\ ~1. ACM, 2004.

\bibitem[Agrawal et~al.(2016)Agrawal, Nair, Abbeel, Malik, and
  Levine]{agrawal2016learning}
Agrawal, Pulkit, Nair, Ashvin~V, Abbeel, Pieter, Malik, Jitendra, and Levine,
  Sergey.
\newblock Learning to poke by poking: Experiential learning of intuitive
  physics.
\newblock In \emph{Advances in Neural Information Processing Systems}, pp.\
  5074--5082, 2016.

\bibitem[Amos \& Kolter(2017)Amos and Kolter]{amos2017optnet}
Amos, Brandon and Kolter, J~Zico.
\newblock Optnet: Differentiable optimization as a layer in neural networks.
\newblock \emph{arXiv preprint arXiv:1703.00443}, 2017.

\bibitem[Andrychowicz et~al.(2017)Andrychowicz, Crow, Ray, Schneider, Fong,
  Welinder, McGrew, Tobin, Abbeel, and Zaremba]{andrychowicz2017hindsight}
Andrychowicz, Marcin, Crow, Dwight, Ray, Alex, Schneider, Jonas, Fong, Rachel,
  Welinder, Peter, McGrew, Bob, Tobin, Josh, Abbeel, Pieter, and Zaremba,
  Wojciech.
\newblock Hindsight experience replay.
\newblock In \emph{Advances in Neural Information Processing Systems}, pp.\
  5055--5065, 2017.

\bibitem[Ba et~al.(2016)Ba, Kiros, and Hinton]{ba2016layer}
Ba, Jimmy~Lei, Kiros, Jamie~Ryan, and Hinton, Geoffrey~E.
\newblock Layer normalization.
\newblock \emph{arXiv preprint arXiv:1607.06450}, 2016.

\bibitem[Bansal et~al.(2017)Bansal, Pachocki, Sidor, Sutskever, and
  Mordatch]{bansal2017emergent}
Bansal, Trapit, Pachocki, Jakub, Sidor, Szymon, Sutskever, Ilya, and Mordatch,
  Igor.
\newblock Emergent complexity via multi-agent competition.
\newblock \emph{arXiv preprint arXiv:1710.03748}, 2017.

\bibitem[Baram et~al.(2017)Baram, Anschel, Caspi, and Mannor]{baram2017end}
Baram, Nir, Anschel, Oron, Caspi, Itai, and Mannor, Shie.
\newblock End-to-end differentiable adversarial imitation learning.
\newblock In \emph{International Conference on Machine Learning}, pp.\
  390--399, 2017.

\bibitem[Brockman et~al.(2016)Brockman, Cheung, Pettersson, Schneider,
  Schulman, Tang, and Zaremba]{brockman2016openai}
Brockman, Greg, Cheung, Vicki, Pettersson, Ludwig, Schneider, Jonas, Schulman,
  John, Tang, Jie, and Zaremba, Wojciech.
\newblock Openai gym.
\newblock \emph{arXiv preprint arXiv:1606.01540}, 2016.

\bibitem[Chen et~al.(2017)Chen, Mishra, Rohaninejad, and
  Abbeel]{chen2017pixelsnail}
Chen, Xi, Mishra, Nikhil, Rohaninejad, Mostafa, and Abbeel, Pieter.
\newblock Pixelsnail: An improved autoregressive generative model.
\newblock \emph{arXiv preprint arXiv:1712.09763}, 2017.

\bibitem[de~Bruin et~al.(2018)de~Bruin, Kober, Tuyls, and Babuska]{8276247}
de~Bruin, T., Kober, J., Tuyls, K., and Babuska, R.
\newblock Integrating state representation learning into deep reinforcement
  learning.
\newblock \emph{IEEE Robotics and Automation Letters}, PP\penalty0
  (99):\penalty0 1--1, 2018.
\newblock \doi{10.1109/LRA.2018.2800101}.

\bibitem[Deguchi \& Takahashi(1999)Deguchi and Takahashi]{deguchi1999image}
Deguchi, Koichiro and Takahashi, Isao.
\newblock Image-based simultaneous control of robot and target object motions
  by direct-image-interpretation method.
\newblock In \emph{Intelligent Robots and Systems, 1999. IROS'99. Proceedings.
  1999 IEEE/RSJ International Conference on}, volume~1, pp.\  375--380. IEEE,
  1999.

\bibitem[Dhariwal et~al.(2017)Dhariwal, Hesse, Klimov, Nichol, Plappert,
  Radford, Schulman, Sidor, and Wu]{baselines}
Dhariwal, Prafulla, Hesse, Christopher, Klimov, Oleg, Nichol, Alex, Plappert,
  Matthias, Radford, Alec, Schulman, John, Sidor, Szymon, and Wu, Yuhuai.
\newblock Openai baselines, 2017.

\bibitem[Donti et~al.(2017)Donti, Kolter, and Amos]{donti2017task}
Donti, Priya, Kolter, J~Zico, and Amos, Brandon.
\newblock Task-based end-to-end model learning in stochastic optimization.
\newblock In \emph{Advances in Neural Information Processing Systems}, pp.\
  5490--5500, 2017.

\bibitem[Farquhar et~al.(2017)Farquhar, Rockt{\"a}schel, Igl, and
  Whiteson]{farquhar2017treeqn}
Farquhar, Gregory, Rockt{\"a}schel, Tim, Igl, Maximilian, and Whiteson, Shimon.
\newblock Treeqn and atreec: Differentiable tree planning for deep
  reinforcement learning.
\newblock \emph{arXiv preprint arXiv:1710.11417}, 2017.

\bibitem[Finn \& Levine(2018)Finn and Levine]{finn2017meta}
Finn, Chelsea and Levine, Sergey.
\newblock Meta-learning and universality: Deep representations and gradient
  descent can approximate any learning algorithm.
\newblock \emph{International Conference on Learning Representations}, 2018.

\bibitem[Finn et~al.(2016{\natexlab{a}})Finn, Levine, and
  Abbeel]{finn2016guided}
Finn, Chelsea, Levine, Sergey, and Abbeel, Pieter.
\newblock Guided cost learning: Deep inverse optimal control via policy
  optimization.
\newblock In \emph{International Conference on Machine Learning}, pp.\  49--58,
  2016{\natexlab{a}}.

\bibitem[Finn et~al.(2016{\natexlab{b}})Finn, Tan, Duan, Darrell, Levine, and
  Abbeel]{finn2016deep}
Finn, Chelsea, Tan, Xin~Yu, Duan, Yan, Darrell, Trevor, Levine, Sergey, and
  Abbeel, Pieter.
\newblock Deep spatial autoencoders for visuomotor learning.
\newblock In \emph{Robotics and Automation (ICRA), 2016 IEEE International
  Conference on}, pp.\  512--519. IEEE, 2016{\natexlab{b}}.

\bibitem[Finn et~al.(2017{\natexlab{a}})Finn, Abbeel, and
  Levine]{finn2017model}
Finn, Chelsea, Abbeel, Pieter, and Levine, Sergey.
\newblock Model-agnostic meta-learning for fast adaptation of deep networks.
\newblock \emph{arXiv preprint arXiv:1703.03400}, 2017{\natexlab{a}}.

\bibitem[Finn et~al.(2017{\natexlab{b}})Finn, Yu, Zhang, Abbeel, and
  Levine]{finn2017one}
Finn, Chelsea, Yu, Tianhe, Zhang, Tianhao, Abbeel, Pieter, and Levine, Sergey.
\newblock One-shot visual imitation learning via meta-learning.
\newblock \emph{arXiv preprint arXiv:1709.04905}, 2017{\natexlab{b}}.

\bibitem[Frans et~al.(2018)Frans, Ho, Chen, Abbeel, and
  Schulman]{frans2018meta}
Frans, Kevin, Ho, Jonathan, Chen, Xi, Abbeel, Pieter, and Schulman, John.
\newblock {META} {LEARNING} {SHARED} {HIERARCHIES}.
\newblock In \emph{International Conference on Learning Representations}, 2018.
\newblock URL \url{https://openreview.net/forum?id=SyX0IeWAW}.

\bibitem[Goodfellow et~al.(2016)Goodfellow, Bengio, and
  Courville]{goodfellow2016deep}
Goodfellow, Ian, Bengio, Yoshua, and Courville, Aaron.
\newblock \emph{Deep learning}, volume~1.
\newblock 2016.

\bibitem[Guez et~al.(2018)Guez, Weber, Antonoglou, Simonyan, Vinyals, Wierstra,
  Munos, and Silver]{guez2018learning}
Guez, Arthur, Weber, Theophane, Antonoglou, Ioannis, Simonyan, Karen, Vinyals,
  Oriol, Wierstra, Daan, Munos, Remi, and Silver, David.
\newblock Learning to search with {MCTS}nets, 2018.
\newblock URL \url{https://openreview.net/forum?id=r1TA9ZbA-}.

\bibitem[Henaff et~al.(2017)Henaff, Whitney, and LeCun]{henaff2017model}
Henaff, Mikael, Whitney, William~F, and LeCun, Yann.
\newblock Model-based planning in discrete action spaces.
\newblock \emph{arXiv preprint arXiv:1705.07177}, 2017.

\bibitem[Higgins et~al.(2017)Higgins, Pal, Rusu, Matthey, Burgess, Pritzel,
  Botvinick, Blundell, and Lerchner]{higgins2017darla}
Higgins, Irina, Pal, Arka, Rusu, Andrei~A, Matthey, Loic, Burgess,
  Christopher~P, Pritzel, Alexander, Botvinick, Matthew, Blundell, Charles, and
  Lerchner, Alexander.
\newblock Darla: Improving zero-shot transfer in reinforcement learning.
\newblock \emph{arXiv preprint arXiv:1707.08475}, 2017.

\bibitem[Ho \& Ermon(2016)Ho and Ermon]{ho2016generative}
Ho, Jonathan and Ermon, Stefano.
\newblock Generative adversarial imitation learning.
\newblock In \emph{Advances in Neural Information Processing Systems}, pp.\
  4565--4573, 2016.

\bibitem[Jonschkowski \& Brock(2015)Jonschkowski and
  Brock]{jonschkowski2015learning}
Jonschkowski, Rico and Brock, Oliver.
\newblock Learning state representations with robotic priors.
\newblock \emph{Autonomous Robots}, 39\penalty0 (3):\penalty0 407--428, 2015.

\bibitem[Jonschkowski et~al.(2017)Jonschkowski, Hafner, Scholz, and
  Riedmiller]{jonschkowski2017pves}
Jonschkowski, Rico, Hafner, Roland, Scholz, Jonathan, and Riedmiller, Martin.
\newblock Pves: Position-velocity encoders for unsupervised learning of
  structured state representations.
\newblock \emph{arXiv preprint arXiv:1705.09805}, 2017.

\bibitem[Kelley(1960)]{Kelley:1960}
Kelley, H.~J.
\newblock Gradient theory of optimal flight paths.
\newblock \emph{ARS Journal}, 30\penalty0 (10):\penalty0 947--954, 1960.

\bibitem[Kingma \& Welling(2013)Kingma and Welling]{kingma2013auto}
Kingma, Diederik~P and Welling, Max.
\newblock Auto-encoding variational bayes.
\newblock \emph{arXiv preprint arXiv:1312.6114}, 2013.

\bibitem[Lange et~al.(2012)Lange, Riedmiller, and
  Voigtlander]{lange2012autonomous}
Lange, Sascha, Riedmiller, Martin, and Voigtlander, Arne.
\newblock Autonomous reinforcement learning on raw visual input data in a real
  world application.
\newblock In \emph{Neural Networks (IJCNN), The 2012 International Joint
  Conference on}, pp.\  1--8. IEEE, 2012.

\bibitem[Levine et~al.(2016)Levine, Finn, Darrell, and Abbeel]{levine2016end}
Levine, Sergey, Finn, Chelsea, Darrell, Trevor, and Abbeel, Pieter.
\newblock End-to-end training of deep visuomotor policies.
\newblock \emph{The Journal of Machine Learning Research}, 17\penalty0
  (1):\penalty0 1334--1373, 2016.

\bibitem[Li et~al.(2017)Li, Song, and Ermon]{li2017inferring}
Li, Yunzhu, Song, Jiaming, and Ermon, Stefano.
\newblock Inferring the latent structure of human decision-making from raw
  visual inputs.
\newblock \emph{arXiv preprint arXiv:1703.08840}, 2017.

\bibitem[Mnih et~al.(2015)Mnih, Kavukcuoglu, Silver, Rusu, Veness, Bellemare,
  Graves, Riedmiller, Fidjeland, Ostrovski, et~al.]{mnih2015human}
Mnih, Volodymyr, Kavukcuoglu, Koray, Silver, David, Rusu, Andrei~A, Veness,
  Joel, Bellemare, Marc~G, Graves, Alex, Riedmiller, Martin, Fidjeland,
  Andreas~K, Ostrovski, Georg, et~al.
\newblock Human-level control through deep reinforcement learning.
\newblock \emph{Nature}, 518\penalty0 (7540):\penalty0 529, 2015.

\bibitem[Nair et~al.(2017)Nair, Chen, Agrawal, Isola, Abbeel, Malik, and
  Levine]{nair2017combining}
Nair, Ashvin, Chen, Dian, Agrawal, Pulkit, Isola, Phillip, Abbeel, Pieter,
  Malik, Jitendra, and Levine, Sergey.
\newblock Combining self-supervised learning and imitation for vision-based
  rope manipulation.
\newblock \emph{arXiv preprint arXiv:1703.02018}, 2017.

\bibitem[Ng \& Russell(2000)Ng and Russell]{ng2000algorithms}
Ng, Andrew~Y and Russell, Stuart.
\newblock Algorithms for inverse reinforcement learning.
\newblock In \emph{in Proc. 17th International Conf. on Machine Learning}.
  Citeseer, 2000.

\bibitem[Ng et~al.(1999)Ng, Harada, and Russell]{ng1999policy}
Ng, Andrew~Y, Harada, Daishi, and Russell, Stuart.
\newblock Policy invariance under reward transformations: Theory and
  application to reward shaping.
\newblock In \emph{ICML}, volume~99, pp.\  278--287, 1999.

\bibitem[Oh et~al.(2017)Oh, Singh, and Lee]{oh2017value}
Oh, Junhyuk, Singh, Satinder, and Lee, Honglak.
\newblock Value prediction network.
\newblock In \emph{Advances in Neural Information Processing Systems}, pp.\
  6120--6130, 2017.

\bibitem[Pathak et~al.(2017)Pathak, Agrawal, Efros, and
  Darrell]{pathakICMl17curiosity}
Pathak, Deepak, Agrawal, Pulkit, Efros, Alexei~A., and Darrell, Trevor.
\newblock Curiosity-driven exploration by self-supervised prediction.
\newblock In \emph{ICML}, 2017.

\bibitem[Pathak* et~al.(2018)Pathak*, Mahmoudieh*, Luo*, Agrawal*, Chen,
  Shentu, Shelhamer, Malik, Efros, and Darrell]{pathak*2018zeroshot}
Pathak*, Deepak, Mahmoudieh*, Parsa, Luo*, Michael, Agrawal*, Pulkit, Chen,
  Dian, Shentu, Fred, Shelhamer, Evan, Malik, Jitendra, Efros, Alexei~A., and
  Darrell, Trevor.
\newblock Zero-shot visual imitation.
\newblock In \emph{International Conference on Learning Representations}, 2018.
\newblock URL \url{https://openreview.net/forum?id=BkisuzWRW}.

\bibitem[Pereira et~al.(2018)Pereira, Fan, An, and Theodorou]{pereira2018mpc}
Pereira, Marcus, Fan, David~D, An, Gabriel~Nakajima, and Theodorou, Evangelos.
\newblock Mpc-inspired neural network policies for sequential decision making.
\newblock \emph{arXiv preprint arXiv:1802.05803}, 2018.

\bibitem[Pinto et~al.(2016)Pinto, Gandhi, Han, Park, and
  Gupta]{pinto2016curious}
Pinto, Lerrel, Gandhi, Dhiraj, Han, Yuanfeng, Park, Yong-Lae, and Gupta,
  Abhinav.
\newblock The curious robot: Learning visual representations via physical
  interactions.
\newblock In \emph{European Conference on Computer Vision}, pp.\  3--18.
  Springer, 2016.

\bibitem[Radford et~al.(2015)Radford, Metz, and
  Chintala]{radford2015unsupervised}
Radford, Alec, Metz, Luke, and Chintala, Soumith.
\newblock Unsupervised representation learning with deep convolutional
  generative adversarial networks.
\newblock \emph{arXiv preprint arXiv:1511.06434}, 2015.

\bibitem[Ramachandran et~al.(2017)Ramachandran, Zoph, and
  Le]{ramachandran2017swish}
Ramachandran, Prajit, Zoph, Barret, and Le, Quoc~V.
\newblock Swish: a self-gated activation function.
\newblock \emph{arXiv preprint arXiv:1710.05941}, 2017.

\bibitem[Sadeghi \& Levine(2016)Sadeghi and Levine]{sadeghi2016cad2rl}
Sadeghi, Fereshteh and Levine, Sergey.
\newblock Cad2rl: Real single-image flight without a single real image.
\newblock \emph{arXiv preprint arXiv:1611.04201}, 2016.

\bibitem[Salimans \& Kingma(2016)Salimans and Kingma]{salimans2016weight}
Salimans, Tim and Kingma, Diederik~P.
\newblock Weight normalization: A simple reparameterization to accelerate
  training of deep neural networks.
\newblock In \emph{Advances in Neural Information Processing Systems}, pp.\
  901--909, 2016.

\bibitem[Schmidhuber(1990)]{Schmidhuber90anon-line}
Schmidhuber, Jürgen.
\newblock An on-line algorithm for dynamic reinforcement learning and planning
  in reactive environments.
\newblock In \emph{In Proc. IEEE/INNS International Joint Conference on Neural
  Networks}, pp.\  253--258. IEEE Press, 1990.

\bibitem[Schneider et~al.(2017)Schneider, Welinder, Ray, Ho, and
  Zaremba]{mujocopy}
Schneider, Jonas, Welinder, Peter, Ray, Alex, Ho, Jonathan, and Zaremba,
  Wojceich.
\newblock Openai mujoco-py, 2017.

\bibitem[Schulman(2016)]{Schulman:EECS-2016-217}
Schulman, John.
\newblock \emph{Optimizing Expectations: From Deep Reinforcement Learning to
  Stochastic Computation Graphs}.
\newblock PhD thesis, EECS Department, University of California, Berkeley, Dec
  2016.
\newblock URL
  \url{http://www2.eecs.berkeley.edu/Pubs/TechRpts/2016/EECS-2016-217.html}.

\bibitem[Schulman et~al.(2017)Schulman, Wolski, Dhariwal, Radford, and
  Klimov]{schulman2017proximal}
Schulman, John, Wolski, Filip, Dhariwal, Prafulla, Radford, Alec, and Klimov,
  Oleg.
\newblock Proximal policy optimization algorithms.
\newblock \emph{arXiv preprint arXiv:1707.06347}, 2017.

\bibitem[Sermanet et~al.(2016)Sermanet, Xu, and
  Levine]{sermanet2016unsupervised}
Sermanet, Pierre, Xu, Kelvin, and Levine, Sergey.
\newblock Unsupervised perceptual rewards for imitation learning.
\newblock \emph{arXiv preprint arXiv:1612.06699}, 2016.

\bibitem[Sermanet et~al.(2017)Sermanet, Lynch, Hsu, and
  Levine]{sermanet2017time}
Sermanet, Pierre, Lynch, Corey, Hsu, Jasmine, and Levine, Sergey.
\newblock Time-contrastive networks: Self-supervised learning from multi-view
  observation.
\newblock \emph{arXiv preprint arXiv:1704.06888}, 2017.

\bibitem[Silver et~al.(2016)Silver, van Hasselt, Hessel, Schaul, Guez, Harley,
  Dulac-Arnold, Reichert, Rabinowitz, Barreto, et~al.]{silver2016predictron}
Silver, David, van Hasselt, Hado, Hessel, Matteo, Schaul, Tom, Guez, Arthur,
  Harley, Tim, Dulac-Arnold, Gabriel, Reichert, David, Rabinowitz, Neil,
  Barreto, Andre, et~al.
\newblock The predictron: End-to-end learning and planning.
\newblock \emph{arXiv preprint arXiv:1612.08810}, 2016.

\bibitem[Sukhbaatar et~al.(2017)Sukhbaatar, Kostrikov, Szlam, and
  Fergus]{sukhbaatar2017intrinsic}
Sukhbaatar, Sainbayar, Kostrikov, Ilya, Szlam, Arthur, and Fergus, Rob.
\newblock Intrinsic motivation and automatic curricula via asymmetric
  self-play.
\newblock \emph{arXiv preprint arXiv:1703.05407}, 2017.

\bibitem[Sutton \& Barto(1998)Sutton and Barto]{suttonbook}
Sutton, Richard~S. and Barto, Andrew~G.
\newblock \emph{Introduction to Reinforcement Learning}.
\newblock MIT Press, Cambridge, MA, USA, 1st edition, 1998.
\newblock ISBN 0262193981.

\bibitem[Tamar et~al.(2016)Tamar, Wu, Thomas, Levine, and
  Abbeel]{tamar2016value}
Tamar, Aviv, Wu, Yi, Thomas, Garrett, Levine, Sergey, and Abbeel, Pieter.
\newblock Value iteration networks.
\newblock In \emph{Advances in Neural Information Processing Systems}, pp.\
  2154--2162, 2016.

\bibitem[Tamar et~al.(2017)Tamar, Thomas, Zhang, Levine, and
  Abbeel]{tamar2017learning}
Tamar, Aviv, Thomas, Garrett, Zhang, Tianhao, Levine, Sergey, and Abbeel,
  Pieter.
\newblock Learning from the hindsight plan—episodic mpc improvement.
\newblock In \emph{Robotics and Automation (ICRA), 2017 IEEE International
  Conference on}, pp.\  336--343. IEEE, 2017.

\bibitem[Tassa et~al.(2012)Tassa, Erez, and Todorov]{tassa2012synthesis}
Tassa, Yuval, Erez, Tom, and Todorov, Emanuel.
\newblock Synthesis and stabilization of complex behaviors through online
  trajectory optimization.
\newblock In \emph{Intelligent Robots and Systems (IROS), 2012 IEEE/RSJ
  International Conference on}, pp.\  4906--4913. IEEE, 2012.

\bibitem[Tobin et~al.(2017)Tobin, Fong, Ray, Schneider, Zaremba, and
  Abbeel]{tobin2017domain}
Tobin, Josh, Fong, Rachel, Ray, Alex, Schneider, Jonas, Zaremba, Wojciech, and
  Abbeel, Pieter.
\newblock Domain randomization for transferring deep neural networks from
  simulation to the real world.
\newblock In \emph{Intelligent Robots and Systems (IROS), 2017 IEEE/RSJ
  International Conference on}, pp.\  23--30. IEEE, 2017.

\bibitem[Todorov et~al.(2012)Todorov, Erez, and Tassa]{todorov2012mujoco}
Todorov, Emanuel, Erez, Tom, and Tassa, Yuval.
\newblock Mujoco: A physics engine for model-based control.
\newblock In \emph{Intelligent Robots and Systems (IROS), 2012 IEEE/RSJ
  International Conference on}, pp.\  5026--5033. IEEE, 2012.

\bibitem[Watter et~al.(2015)Watter, Springenberg, Boedecker, and
  Riedmiller]{watter2015embed}
Watter, Manuel, Springenberg, Jost, Boedecker, Joschka, and Riedmiller, Martin.
\newblock Embed to control: A locally linear latent dynamics model for control
  from raw images.
\newblock In \emph{Advances in neural information processing systems}, pp.\
  2746--2754, 2015.

\bibitem[Weber et~al.(2017)Weber, Racani{\`e}re, Reichert, Buesing, Guez,
  Rezende, Badia, Vinyals, Heess, Li, et~al.]{weber2017imagination}
Weber, Th{\'e}ophane, Racani{\`e}re, S{\'e}bastien, Reichert, David~P, Buesing,
  Lars, Guez, Arthur, Rezende, Danilo~Jimenez, Badia, Adria~Puigdom{\`e}nech,
  Vinyals, Oriol, Heess, Nicolas, Li, Yujia, et~al.
\newblock Imagination-augmented agents for deep reinforcement learning.
\newblock \emph{arXiv preprint arXiv:1707.06203}, 2017.

\bibitem[Wulfmeier et~al.(2016)Wulfmeier, Rao, and
  Posner]{wulfmeier2016incorporating}
Wulfmeier, Markus, Rao, Dushyant, and Posner, Ingmar.
\newblock Incorporating human domain knowledge into large scale cost function
  learning.
\newblock \emph{arXiv preprint arXiv:1612.04318}, 2016.

\end{thebibliography}
\bibliographystyle{icml2018}

\appendix


\twocolumn[
\icmltitle{Universal Planning Networks - Supplementary}






\vskip 0.3in
]




\section{UPN Architecture details}

We overview the details of the architecture to aid in reproducing the work. 

\subsection{Input}

Our current and goal observations $(o_t, o_g)$ are 84 $\times$ 84 $\times$ 3 RGB images. Our objective is to re-use the learned representations (encodings) of pixel inputs. Thus, we kept things simple and did not apply any standardization techniques like mean subtraction or input batch normalization / layer normalization because it's not clear  how such input transformations would transfer to new tasks. However, we scale the input pixels by $\frac{1}{255}$.

\subsection{Encoder $f_\phi$}

Below, we describe the sequence of convolutional and feed-forward layers. These operations together make our encoder $f_\phi$ described in the UPN architecture. Note that we use the {\texttt{swish}} nonlinearity that was proposed by \citet{ramachandran2017swish}. We found {\texttt{swish}} to work better than {\texttt{relu}} for all our experiments. For more clarity, $\textrm{swish}(x) = \textrm{sigmoid}(x)*x$. The code snippet below returns 128-dimensional vectors for $x_t$ and $x_g$.

\begin{lstlisting}
import tensorflow.contrib.layers as layers
# o: input pixel observation, 
# x: encoded representation
# Encoder transformation o ---> x, 
# example: o_t ---> x_t, o_g ---> x_g
h = layers.convolution2d(o, num_outputs=32, kernel_size=8, stride=4, padding='VALID', activation_fn=None)
h = tf.nn.sigmoid(h)*h
h = layers.convolution2d(h, num_outputs=64, kernel_size=4, stride=2, padding='VALID', activation_fn=None)
h = tf.nn.sigmoid(h)*h
h = layers.convolution2d(h, num_outputs=64, kernel_size=3, stride=1, padding='VALID', activation_fn=None)
h = tf.nn.sigmoid(h)*h
h = layers.convolution2d(h, num_outputs=16, kernel_size=2, stride=1, padding='VALID', activation_fn=None)
h = tf.nn.sigmoid(h)*h
for _ in range(2):
    h = layers.fully_connected(h, num_outputs=128, activation_fn=None)
    h = layers.layer_norm(h, center=True, scale=True)
    h = tf.nn.sigmoid(h)*h
\end{lstlisting}

\subsection{Bias Transformation and Embodiment Information}

In our experiments, we use torque-controlled agents. It is an ill-defined problem to map purely from pixels to motor torques without information about the joint velocities. Though it may be possible to recover the joint angles and end-effector positions from pixels, recovering the velocities requires multiple frames from the past, thereby making the problem of mapping from current image observation to motor torques a partially observed problem. To get rid of this problem, researchers in the robotic learning community have typically provided the joint angles and velocities as additional inputs that are concatenated with the encoded features of the image observation~\cite{levine2016end, finn2017one}. We follow the same technique for our paper. Note that even though providing the robot arm configuration is helpful for planning motor torques, the agent also needs to rely on learning a good perceptual representation and reward function, since the joint angles and velocities do not contain any information about the goal or task at hand. 

In addition to the above, we also borrow another trick called the bias transformation~\cite{finn2017one} to ensure stable gradients in inner optimization loop of complex computation graphs. The bias transformation units are free variables that increase the expressivity of the gradient without adding expressivity to the network, allowing for more control of the gradient. See \citet{finn2017one} for a more detailed explanation of bias transformation and its usage in model-agnostic meta-learning (MAML) \cite{finn2017model}.

To summarize, our inner loop planner tries to identify a sequence of actions that are optimal for reaching $o_g$ from $o_t$. Having transformed $o_t$ and $o_g$ to $x_t$ and $x_g$ respectively, we seek $\hat{a}_{t:t+T}$ that when rolled forward from $x_t$, produces a latent state close to $x_g$. This forward unrolling takes in additional inputs $q_t$ and $b$, where $q_t$ provides the sensorimotor embodiment information, and $b$ denotes the bias transformation variable. We re-project $x_t$ to a 128-dimensional space since $x_g$ is still 128-dimensional. We present the code relevant to these tricks below:

\begin{lstlisting}
# xt = encoded version of o_t
# qt = joint angles and velocities at time step t
# bt_num_units = number of bias transform dimensions
xt = tf.concat([xt, qt], axis=1)
bt_num_units = 20 
bias_transform = tf.get_variable(
    'bias_transform',
    [1,bt_num_units],
    initializer=tf.constant_initializer(0.1))
bias_transform = tf.tile(bias_transform, 
    multiples=tf.stack([tf.shape(xt)[0], 1]))
xt = tf.concat([xt, bias_transform], 1)
# Project back to 128 dimensions
xt = layers.fully_connected(x, num_outputs=128, activation_fn=None)
xt = layers.layer_norm(out, center=True, scale=True)
xt = tf.nn.sigmoid(xt)*xt


\end{lstlisting}

\subsection{Action Encoder}

We encode the actions which are in turn used by the forward dynamics model in the latent space. Our action encoder is as follows:

\begin{lstlisting}
# plan - current sequence of actions, batch_size x horizon x act_dim
plan = tf.reshape(plan, [-1, act_dim])
plan = layers.fully_connected(plan, num_outputs=64, activation_fn=None)
plan = layers.layer_norm(plan, center=True, scale=True)
plan = tf.nn.sigmoid(plan)*plan
plan = tf.reshape(plan, [-1, horizon, 64])
\end{lstlisting}

\subsection{Planning by Gradient Descent}

Our planner rolls forward a dynamics model in the latent space of the encoder using the current estimate of the optimal plan. Our dynamics model prediction for one-step is best summarized as a rough skeleton code below:

\subsubsection{Dynamics in Latent Space}

\begin{lstlisting}
# One-step dynamics in latent space
# Note: curr_state, next_state, action are in the latent spaces.
next_state = layers.fully_connected(
    tf.concat([curr_state, action]),
    out_dim=128,
    nonlinearity=None)
next_state = layers.layer_norm(next_state, center=True, scale=True)
next_state = tf.nn.sigmoid(next_state)*next_state

\end{lstlisting}

\subsubsection{Planning Updates}

We present {\it pseudo-code} below for better understanding of the planning process, which uses multiple gradient descent steps on a Huber Loss, with respect to the actions. 

\begin{lstlisting}

# g_theta: dynamics, f_phi: encoder
# ot: current obs, og: goal obs, 
# plan: sequence of actions
# qt: joint angles and velocities of robot
# b: bias transformation variable
# T: horizon 

# Encode observations to latent
xg = f_phi(og) # 128 dimensional
xt = f_phi(ot) # 128 dimensional
xt_joint = tf.concat([xt, qt,b],1) 
xg_pred = fully_connected(xt_joint) 
# 128 dimensional 

# Encode plan to latent
latent_plan = fully_connected(plan) 
# 64 dimensional

for update_step in range(num_updates):

    # Roll out dynamics in latent
    for timestep in range(T):
        xg_pred = g_theta(
        xg_pred, latent_plan[timestep])

    # Compute plan error
    error = tf.losses.huber_loss(
    xg, xg_pred, huber_delta) 
    
    # Compute plan gradients
    plan_grad = tf.gradients(error, plan)
    
    # Improve plan using gradient descent
    plan = plan - step_size * plan_grad


\end{lstlisting}

\subsubsection{Hyperparameter details}

\begin{table}[h]
\footnotesize
\caption{Hyperparameters for planning module}
\label{gdp-table}
\vskip 0.15in
\begin{center}
\begin{small}
\begin{sc}
\begin{tabular}{lcccr}
\toprule
 & Hyperparameter & Value\\
\midrule
1 & Number of plan updates & 40 \\
2 & Gradient Clip Value & 25 \\
3 & Update Step Size(s) & 0.5, 0.25 \\
4 & Huber Delta & 0.85 \\
5 & Planning Horizon (2D Point Robot) & 50  \\ 
6 & Planning Horizon (3-Link Reacher) & 100  \\ 
7 & Planning Horizon (3D Poking) & 100  \\ 
\bottomrule
\end{tabular}
\end{sc}
\end{small}
\end{center}
\vskip -0.1in
\end{table}

{\textbf{Gradient Clipping:}} We use a simple gradient update to the actions (though exploring momentum based updates like Adam could be an interesting direction for future work). However, we clip the gradients using Tensorflow's {\texttt{clip-by-value}} with limits -25 and 25. Further, the step sizes are important for getting the model to train well. 

{\textbf{Huber Loss:}} This trick was, by far, the {\emph{most important}}
to getting the model to train with well behaved inner loop losses and gradients. We also tuned the {\texttt{delta}} parameter in the huber loss ({\texttt{delta=1}} is standard). In our experiments, having a {\texttt{delta=0.85}} turned out to be optimal. In the absence of the Huber Loss, the inner loop planning error increased over training agnostic of whether gradient clipping was used or not. It is also worth noting that \citet{sermanet2017time} use a Huber-style loss on their features to describe reward functions, though the feature space wasn't optimized to provide such metrics at training time.

{\textbf{Step Size:}} We take an aggressive step size of 0.5 for the first update, and step sizes of 0.25 for the remaining 39 updates during training. At test-time, we preserve the same constants. Exploring other schemes (for example, geometric decaying) is left for future work.

{\textbf{Horizon:}} The planning horizon listed in Table \ref{gdp-table} is the maximum number of time steps to roll out planning across time. Thus, we do not always roll out the dynamics across time for the specified horizon for every pair of $(o_t, o_g)$. In practice, it is more data-efficient to extract many sub-sequences from a single trajectory by picking different $(o_{t_1}, o_{t_2})$ where $t_2 > t_1$ and roll out the planning with a dynamic horizon. The implementation thus resembles a \href{https://www.tensorflow.org/api_docs/python/tf/nn/dynamic_rnn}{dynamic RNN}. Further, it is useful to incentivize sampling of shorter horizons over the longer ones early in training, drawing inspiration from curriculum learning. 
We describe the specific details of the curriculum in the next subsection.

\subsection{Outer Loop Loss and Training}

Our outer objective is the standard behavior cloning loss. The updated plan from the inner loop is optimized against the optimal plan we have from our demonstrations. We use a mean squared error loss function because our actions are continuous. This corresponds to a maximum likelihood estimate on a Gaussian distribution with the identity matrix as the covariance. 

In Table \ref{upn-table}, we describe the hyperparameters associated with the outer loop training:

\begin{table}[h]
\footnotesize
\caption{Hyperparameters for training}
\label{upn-table}
\vskip 0.15in
\begin{center}
\begin{small}
\begin{sc}
\begin{tabular}{lcccr}
\toprule
 & Hyperparameter & Value\\
\midrule
1 & Batch Size & 128 \\
2 & Update Rule & Adam \\
3 & Learning Rate & 3e-4 \\
5 & Number of Batch Updates & 1e6 \\
6 & Validation Frequency & 1e4  \\ 
7 & Batch Sampling & Curriculum  \\ 
8 & Input Pixel Scaling & 1/255 \\
\bottomrule
\end{tabular}
\end{sc}
\end{small}
\end{center}
\vskip -0.1in
\end{table}

{\textbf{Larger Batch Size:}} Using a large batch size (128) was crucial. It may be possible that using larger batch sizes (example, 256 and more) with smaller Adam step sizes could be better. We leave that analysis for future work.

{\textbf{Validation Frequency:}} Every 10000 batch updates, we validate the network using a hold-out set from the training data as is typical in supervised learning experiments. We then pick the best network over the course of a million mini-batch updates. The wall clock time for this training was around 1.5 days on an NVIDIA Titan X. Better wall clock times are certainly achievable using multiple GPUs and larger batch sizes.

{\textbf{Batch Sampling:}} As pointed out in the previous section, it is important to bring in the aspect of curriculum training. The credit assignment over longer planning horizons may be easier if the model has learned on shorter horizons. However, pre-training and fine-tuning is expensive and needs more hyper-parameters. Instead, we just skew the sampling distribution to pick the shorter horizons early on during the training, and uniformly later on. For the skewed sampling, we adopt the Poisson distribution over the horizon range. The Poisson sampling was used for the first 300000 mini-batch updates. This trick had a {\it substantial effect} on good training curves.  

\subsection{Things that did not work}

1.   {\textbf{Weight normalization}}:  The motivation for using the Huber Loss was to stop the inner loop gradients, latent features, and planning error from becoming too large. However, the first trick we tried to get rid of the problem was to enforce weight normalization \cite{salimans2016weight} on the features. Interestingly, \citet{farquhar2017treeqn} also point out the same issue when trying to backpropogate through tree-based planning structures inside the computation graph. Though the weight normalization tricked turned out to work for them, we did not have success with it for our case.  

2.   {\textbf{Larger batch size training (256 and 512):}} We weren't successful in scaling up the training and observing gains out of it. The performance on training with batch sizes of 256 was similar to that of 128, while 512 resulted in unstable training.

3.  {\textbf{Batch normalization on inputs:}}  Similar to general observations that batch normalization isn't useful yet for imitation or reinforcement learning problems, we weren't able to observe benefits. In fact, using batch normalization hurt the training because of correlations between observations within small periods of time.

4.  {\textbf{Layer Normalization on convolution layers:}} Similar to observations in \citet{ba2016layer}, layer normalization in convolution layers hurt the training of UPNs. Therefore, we only use layer norm in our recurrent dynamics and other fully connected layers. 

\section{Action-selection with Model Predictive Control (MPC)}

\begin{figure}[H]
\centering
\includegraphics[width=0.3\textwidth]{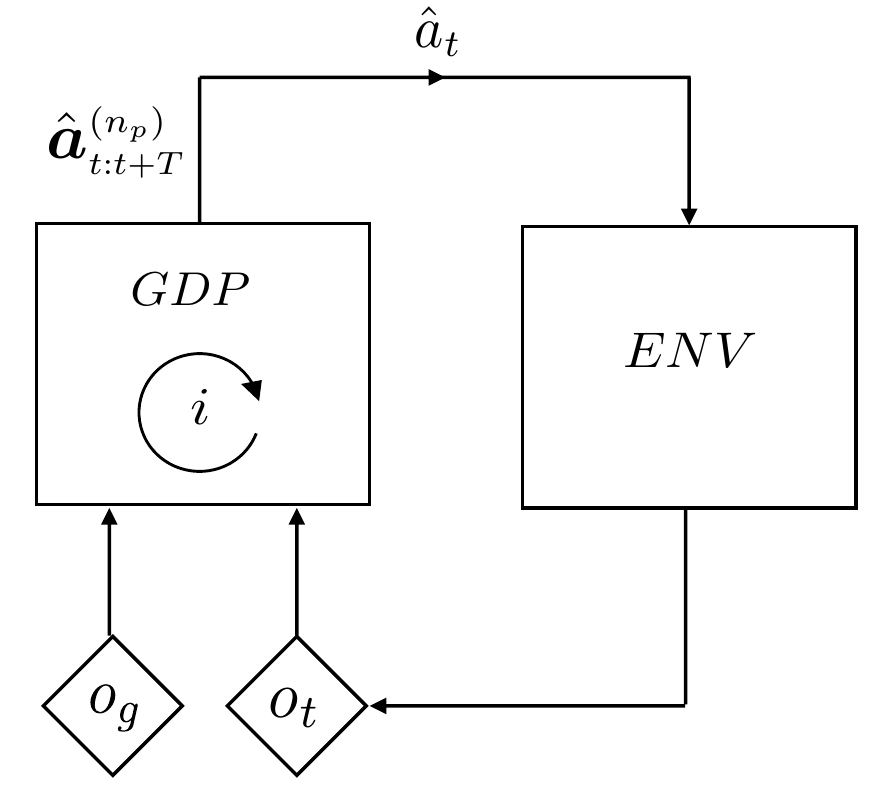} 
\caption[width=0.7\textwidth]{The agent plans a trajectory aiming to reach $o_{g}$ from $o_t$, but only executes the first action in the inferred plan, before replanning.}
\label{fig:mpc}
\end{figure}

In our imitation learning experiments, our action selection procedure uses MPC at test time. The pipeline for UPN is described in Figure~\ref{fig:mpc} for better clarity. The pipeline for AIL is similar to UPN (plan but execute only the first action); while for RIL, we predict only one action and therefore execute it without any notion of MPC. 

\section{Demo collection}
\label{democollect}

We designed the 2D point robot and 3-link reacher tasks to necessitate planning. This makes collecting demonstrations for these environments tricky because off-the-shelf trajectory optimization and model-free algorithms require a well-defined cost function. Such cost functions are not straight-forward to engineer in this case due to the presence of obstacles. We adopt the hindsight experience replay (HER) technique \cite{andrychowicz2017hindsight}, which implicitly learns to shape a cost function by optimizing for task success instead of extrinsic rewards. Navigation and reaching around obstacles naturally fall into the category of path-planning problems which have a nice continual progression of goals in the configurational space associated with the tasks. Thus, HER is suitable for solving such tasks and can provide the demonstrations that we seek. This. however, comes with a couple of issues: (1) HER may not be able to solve the tasks perfectly, especially in the case of longer horizons; (2) The policies learned using HER are not general, i.e., HER would overfit or master a policy for a specific configuration of obstacles when you work in the configurational space (instead of pixel space). To deal with the second issue, we train independent HER policies for each obstacle configuration in our training distribution. However, the learned policies may not be accurate. To deal with this issue (1), we introduce a technique that we call as {\textbf{hindsight rendering}}. Given a goal to reach, we roll out the parametric expert tasked with achieving the goal. If the rollout was successful, it is added to the list of demonstrations. Else, we re-render the same rollout, now with the goal being at the final position of the agent (or that of its end-effector). This becomes a meaningful trajectory in most cases and hence, can be added to the list of demonstrations. 

\section{Imitation Learning Comparisons : Description and architecture details}

For our comparisons, the network topology is described below:

\subsection{Reactive Imitation Learner (RIL)}
Here, $f_\phi$ is a conventional feed-forward architecture with a convolutional encoder and a fully connected layer $m$, to decode the action from the latent representation. Refer to Figure \ref{fig:mlp_baseline}.
\vspace{-2mm}
$$f_{\phi}(o_t, o_g) = x_t$$
$$m_{}(x_t) = \hat{a}_{t}$$
\vspace{-6mm}

\subsection{Autoregressive Imitation Learner (AIL)}
\begin{figure}[h]
\centering     
\subfigure[Reactive imitation learner]{\label{fig:mlp_baseline}\includegraphics[width=0.5\linewidth]{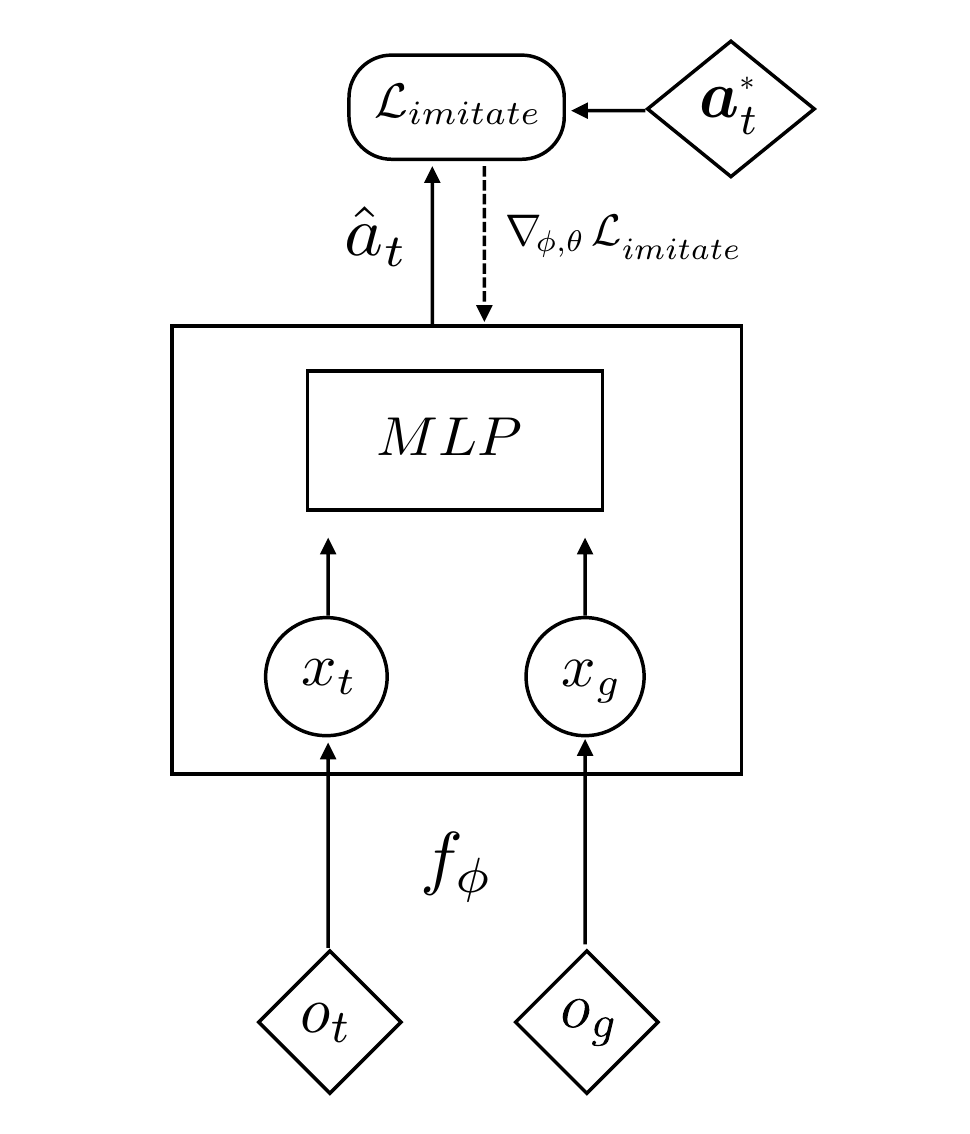}}%

\subfigure[Auto-regressive imitation learner.]{\label{fig:rnn_baseline}\includegraphics[width=\linewidth]{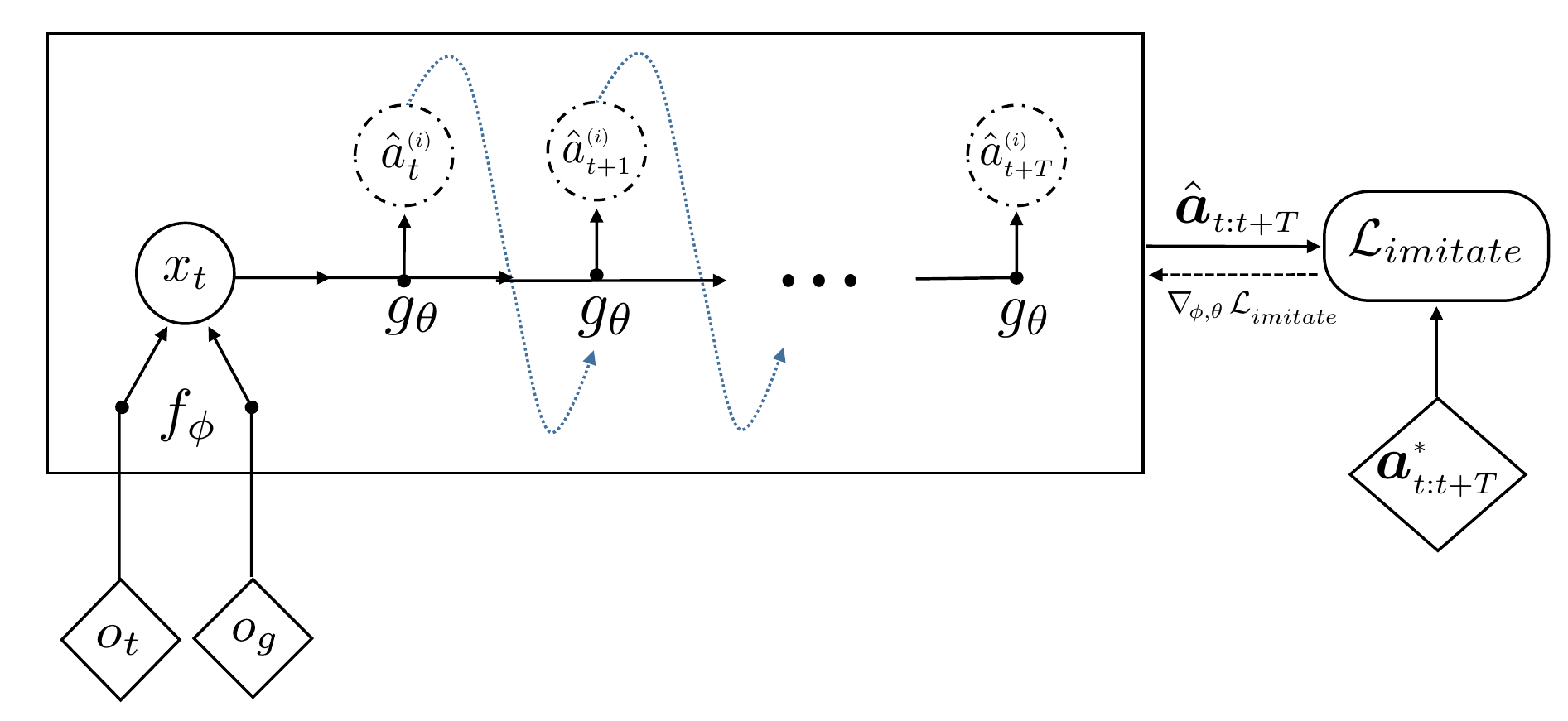}}
\caption{Baseline systems.}
\end{figure}

This model auto-regresses a sequence of actions conditioned on a latent representation. $f_\phi$ is the convolutional encoder which produces the initial hidden state, $g$ is a recurrent cell, and $m$ is a fully connected layer which decodes the action from the recurrent latent representation. The initial hidden state is initialized with the latent representation produced by $f$. Refer to Figure \ref{fig:rnn_baseline}.
$$f_{\phi}(o_t, o_g) = x_t$$
\vspace{-2mm}
$$g_{\theta}(x_t, h_{t-1}) = h_{t} $$
\vspace{-2mm}
$$m_{}(h_t) = \hat{a}_{t} $$

\subsection{Details relevant to RIL and AIL training}

{\textbf{RIL:}} We retain the $f_\phi$ described in the UPN architecture for fair comparisons. Similar to how UPN uses joint angles and velocities for torque control outputs, we fuse the convolutional encoding with the joint information $s_t$ using 2 hidden layers of dimension 128. This is finally followed by an output later corresponding to the number of actions needed for the task. 
 
{\textbf{AIL:}} Similar to RIL, the $f_\phi$ retention from UPN and joint encoding concatenation exist in AIL for fair comparisons. We unroll a recurrent computation over the planning horizon that corresponds to the number of time steps between $o_t$ and $o_g$ in the trajectory. The recurrent computation is a traditional vanilla RNN. We did not observe any benefits from using more sophisticated cells like GRU. However, we leave it for future work to explore temporal convolutions with the caveat that any benefit from the specific recurrent computation used in more sophisticated cells can also be adopted for improving UPN. The recurrent dynamics are of similar expressive power as that of the UPN (128 dimensions for the hidden state). Similar to UPN, we adopt the same skewed sampling of the batches to encourage curriculum learning while training AIL. The actions are decoded from the recurrent hidden states using a single fully connected layer mapping to the action dimension of the task.

The hyperparameters that further clarify the RIL and AIL training details are described in Table \ref{rail-table}. To stay consistent, we have the same train, validation and test splits for UPN, RIL and AIL training. 

\begin{table}[ht]
\footnotesize
\caption{Hyperparameters for RIL and AIL}
\label{rail-table}
\vskip 0.05in
\begin{center}
\begin{small}
\begin{sc}
\begin{tabular}{lcccr}
\toprule
 & Hyperparameter & Value\\
\midrule
1 & Batch Size & 128 \\
2 & Update Rule & Adam \\
3 & Learning Rate & 3e-4 \\
5 & Number of Batch Updates & 1e6 \\
6 & Validation Frequency & 1e4  \\ 
7 & Batch Sampling (AIL) & Curriculum  \\ 
8 & Batch Sampling (RIL) & Uniform \\
9 & Input Pixel Scaling & 1/255 \\
\bottomrule
\end{tabular}
\end{sc}
\end{small}
\end{center}
\vskip -0.1in
\end{table}

\section{Imitation Learning Benchmarks Experimental Setup}

We clarify two ambiguous details that were left out in the main paper:

{\textbf{Expert Performance:}} 
Note that the expert performance is not optimal, especially in the reacher VOVG setting. To be clear, this does \emph{not} mean that the representations of UPN/RIL/AIL are learned from imperfect demonstrations.
As explained in Appendix~\ref{democollect}, the demos are meaningfully correct executions because of hindsight rendering. Thus, the parametric expert is not perfect in terms of achieving the different goals used for training and evaluation. Hence, the reported average success of the expert is lower than 100\%, particularly on the harder tasks like Reacher VOVG.

{\textbf{Notion of success:}} It is important we clarify the notion of what a successful trajectory (for a given goal) is. For both the point robot and the reacher (FOVG and VOVG) settings, a rollout is deemed successful if the end-effector (read as center of mass in the 2D point-robot case) is within 0.05 meters of the goal location. 


Table \ref{il-table} contains the task-specific hyperparameters: maximum number of time steps to roll out the plan (planning horizon), number of demonstrations, and the number of actions. Both the tasks use a success threshold of 0.05 meters.  

\begin{table}[h]
\footnotesize
\caption{Hyperparameters for Imitation Learning Tasks}
\label{il-table}
\vskip 0.15in
\begin{center}
\begin{small}
\begin{sc}
\begin{tabular}{lcccr}
\toprule
Task & Type & Max Horizon & Demos & Actions  \\
\midrule
Point Robot & FOVG  & 50 & 10000 & 2 \\ 
Point Robot & VOVG & 50 & 20000 & 2 \\
Reacher & FOVG & 100 & 20000 & 3 \\ 
Reacher & VOVG & 100 & 40000 & 3 \\ 
\bottomrule
\end{tabular}
\end{sc}
\end{small}
\end{center}
\vskip -0.1in
\end{table}

\section{PPO Parameters}

Table \ref{ppo-table-1} contains the hyperparameters in the RL experiments that we ran for the reacher morphology transfer, 2D point robot to ant transfer (shorter horizons), and the poking to pushing transfer.

\begin{table}[h]
\footnotesize
\caption{PPO hyperparameters used for simpler RL experiments}
\label{ppo-table-1}
\vskip 0.15in
\begin{center}
\begin{small}
\begin{sc}
\begin{tabular}{lcccr}
\toprule
 & Hyperparameter & Value\\
\midrule
1 & Timesteps per actorbatch & 4096 \\
2 & Adam Stepsize & 5e-5 \\
3 & Number of simulation steps & 1e8 \\
4 & Number of epochs & 1 \\
5 & GAE parameter ($\lambda$) & 0.95 \\
6 & Minibatch size & 256 \\ 
\bottomrule
\end{tabular}
\end{sc}
\end{small}
\end{center}
\vskip -0.1in
\end{table}

Table \ref{ppo-table-2} contains the hyperparameters for the long-horizon ant navigation experiment.

\begin{table}[h]
\footnotesize
\caption{PPO hyperparameters used for long-horizon ant RL experiment}
\label{ppo-table-2}
\vskip 0.15in
\begin{center}
\begin{small}
\begin{sc}
\begin{tabular}{lcccr}
\toprule
 & Hyperparameter & Value\\
\midrule
1 & Timesteps per actorbatch & 65536 \\
2 & Adam Stepsize & 5e-5 \\
3 & Number of simulation steps & 1e10 \\
4 & Number of epochs & 3 \\
5 & GAE parameter ($\lambda$) & 0.95 \\
6 & Minibatch size & 1024 \\ 
\bottomrule
\end{tabular}
\end{sc}
\end{small}
\end{center}
\vskip -0.1in
\end{table}

Table \ref{ppo-table-3} contains the hyperparameters for the humanoid experiment.

\begin{table}[h]
\footnotesize
\caption{PPO hyperparameters used for humanoid RL experiment}
\label{ppo-table-3}
\vskip 0.15in
\begin{center}
\begin{small}
\begin{sc}
\begin{tabular}{lcccr}
\toprule
& Hyperparameter & Value\\
\midrule
1 & Timesteps per actorbatch & 16384 \\
2 & Adam Stepsize & 5e-5 \\
3 & Number of simulation steps & 1e9 \\
4 & Number of epochs & 3 \\
5 & GAE parameter ($\lambda$) & 0.95 \\
6 & Minibatch size & 512 \\ 
\bottomrule
\end{tabular}
\end{sc}
\end{small}
\end{center}
\vskip -0.1in
\end{table}

Table \ref{ppo-table-4} contains specific details about the environments, such as horizon and discount factor.

\begin{table}[h]
\footnotesize
\caption{Environment Specific Details}
\label{ppo-table-4}
\vskip 0.15in
\begin{center}
\begin{small}
\begin{sc}
\begin{tabular}{lcccr}
\toprule
& Environment & Horizon  & Discount  \\
\midrule
1 & Ant (simple) & 1500 &   0.999 \\
2 & 5-link reacher  & 250  & 0.995\\
3 & Pushing & 200  & 0.99 \\
4 & Ant (hard) & 8000   & 0.999875\\
5 & Humanoid & 4000 & 0.99975\\

\bottomrule
\end{tabular}
\end{sc}
\end{small}
\end{center}
\vskip -0.1in
\end{table}

{\textbf{Network architecture:}} For all the experiments, we used fully-connected neural networks with 3 hidden layers of 128 dimensions each, and {\texttt{tanh}} nonlinearities for both the policy and the value function. The policy is Gaussian with a state-independent diagonal covariance that is learned. Further, we run each experiment with 3 random seeds and average the results.

\section{Reinforcement Learning Benchmarks Experimental Setup}

\subsection{Comparison Methods}

We evaluate the learned feature spaces based on whether distance metrics described on them can be used as rewards for reinforcement learning, particularly on new problems where all we have are image targets and no demonstrations or extrinsic rewards. A representation which can do this effectively is what we call a {\it generalizable} and {\it plannable} representation. Since we already compare to AIL and RIL in our imitation learning experiments, we continue to analyze the feature spaces learned by those models (AIL and RIL) for the reinforcement learning scenario as well. However, in addition to the feature spaces learned through supervised learning, we also run an experiment to train a variational auto-encoder (VAE) \cite{kingma2013auto} . This VAE is trained with all the images in our demonstration trajectories (similar to \citet{higgins2017darla}). We learn the decoder with a mean-squared-error loss.
We pull out the representation ($z$) of the VAE after sufficient training and refer to it the VAE feature space. 

\subsection{VAE Architecture Details}

For a fair comparison, we preserve the convolutional encoder structure $f_\phi$ that was used in the UPN, RIL and AIL architectures. The gaussian posterior is 128 dimensional, and is decoded back to the pixel space by reversing the convolutional encodings through deconvolutional layers. We scale the input pixels by $\frac{1}{255}$ and decode the outputs through a sigmoid nonlinearity on the predicted pixels to stay in the $[0,1]$ range. However, here are a three points to be made:

1. It is possible that using other nonlinearities (instead of {\texttt{swish}}) are better for training a VAE. The {\texttt{swish}} nonlinearity was discovered through architecture search on supervised learning benchmarks. It is possible that {\texttt{swish}} doesn't apply broadly to unsupervised learning objectives and nonlinearities such as {\texttt{leaky-relu}} \cite{radford2015unsupervised} are better. 

2. The traditional VAE is not the best generative model among the class of models that parametrize the likelihood of the pixels. Experimenting with the latest techniques such as autoregressive decoders with temporal convolutions \cite{chen2017pixelsnail} is necessary and we leave it for future work. 

3. More expressive architectures  helpful / better in terms of the metric used to judge the representations, but they are very likely not likely to change the conclusion of the paper since there is no notion of agency or planning embedded in the latent code of pure pixel prediction objectives.

\subsection{Reacher Morphology Transfer}

The physical dynamics and action spaces of the 3 and 4 link reachers are different. Our objective was to acquire a common representation across the demonstrations from both these robots. To get around this problem and still acquire a shared representation, we train a UPN on these tasks with different $g_\theta$ but shared $f_\phi$. We use 20000 demonstrations each from the 3-link and 4-link reachers. For comparing to our other feature spaces, we train them with shared encoders and branched individual output heads corresponding to each task. This is referred to as Multi-Head in the legend of the corresponding plot (Figure \ref{fig:reachertransferresult}) in the main paper. The VAE training does not consider actions. Thus, we train the VAE as usual, but with an aggregated dataset containing the demonstration snapshots from both the tasks. The success threshold is being within 0.05 meters of the goal.

\subsection{Point Robot to Ant Transfer}

We collected 40000 trajectories with 40 randomized point robot appearances. These trajectories are used to train UPN, RIL, AIL and VAE for studying the point to ant (locomotion around obstacles) transfer.  The reason for this randomization was to ensure that the learned UPN representation can meaningfully interpret a creature like an ant which has extra limbs apart from the central torso (that more or less resembles a point robot when seen top down). The appearance randomization adds planar non-actuated limbs to a point robot. This biases the architectures to process the center of the robot and allows them to work for a top-down snapshot of an ant at test time. The success threshold for the ant task is being within 0.5 meters of the goal.

\subsection{Learning to Push from Poking Trajectories}

It is hard to concretely define the poking task. As long as the object is nudged to a sufficiently different pose, it is considered a successful poke. The question we asked is: If we collected a dataset of random but successful pokes, can the robot learn to associate meaningful patterns such as end-effector being close to the object to be able to displace it, the direction and impulse of the movement for a corresponding pose change, etc. If this were possible and were captured in a latent space, it is natural to expect that latent space to be able to provide shaped metric based rewards for a sophisticated skill such as pushing. Unlike the other kinds of transfer studied in this paper, this experiment is about extrapolation of skills on the same robot. We collected a dataset of 20000 pokes and trained the UPN on these trajectories. The success threshold is for the object to be within 0.08 meters of the goal. The goal location in the pushing experiment and the randomization is over different initial positions of the objects (target and distractor). Thus, as described in subsection \ref{rlexplanation} in the main paper, we don't feed in $f_\phi(o_g)$ to the RL policy at test time.

\section{Camera viewpoint for each task}

For each task, we present below the viewpoint provided to the neural network architectures. 

\subsection{2D point robot task}

\begin{figure}[H]
    \centering
    \includegraphics[width = 0.2 \textwidth]{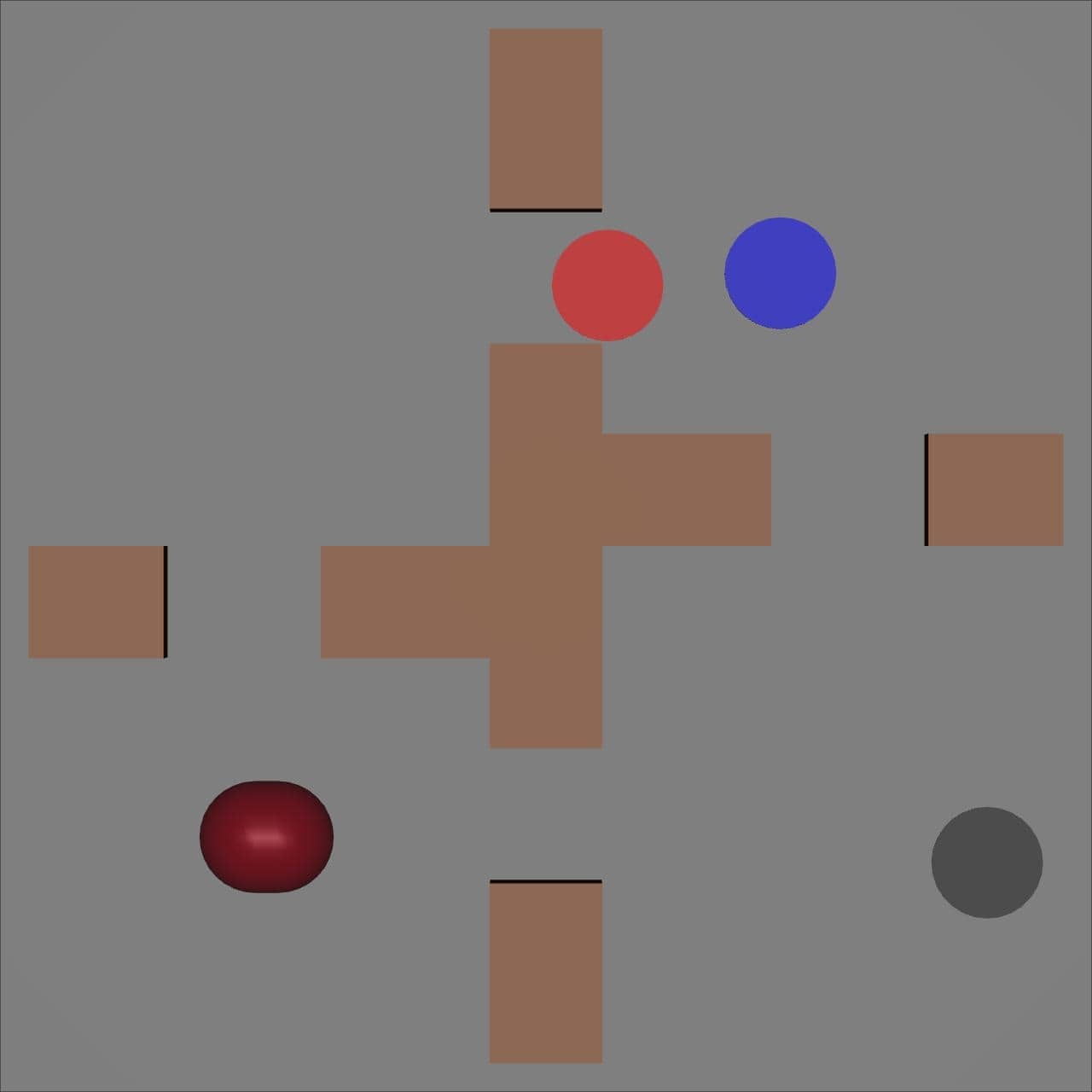}
    \caption{2D Point Robot navigating around obstacles}
    \label{fig:2dtopdown}
\end{figure}

\subsection{Ant around obstacles task}

\begin{figure}[H]
    \centering
    \includegraphics[width = 0.2 \textwidth]{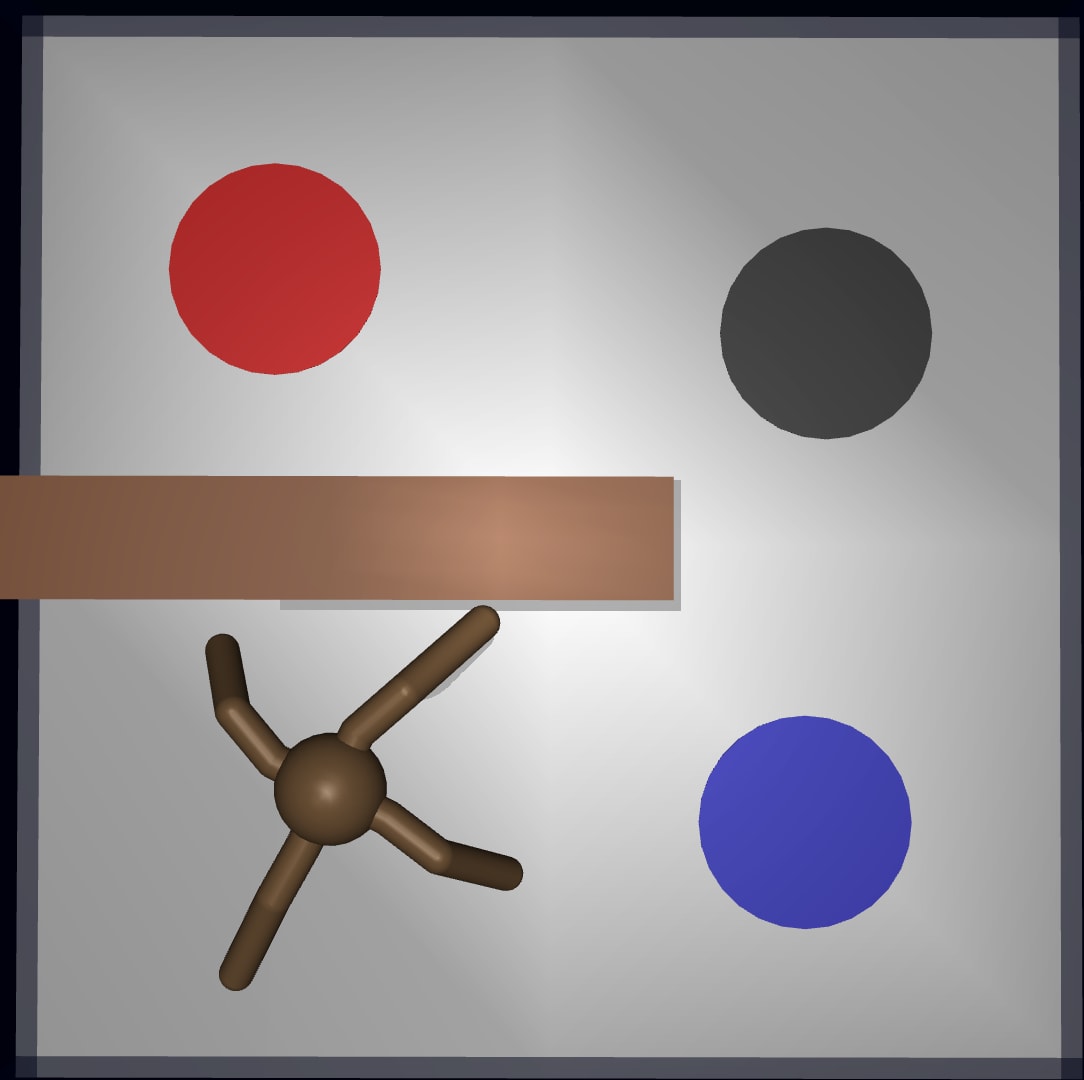}
    \caption{Ant robot navigatng around obstacles}
    \label{fig:anttopdown}
\end{figure}

\subsection{Reacher around obstacles task}
\begin{figure}[H]
    \centering
    \includegraphics[width = 0.2 \textwidth]{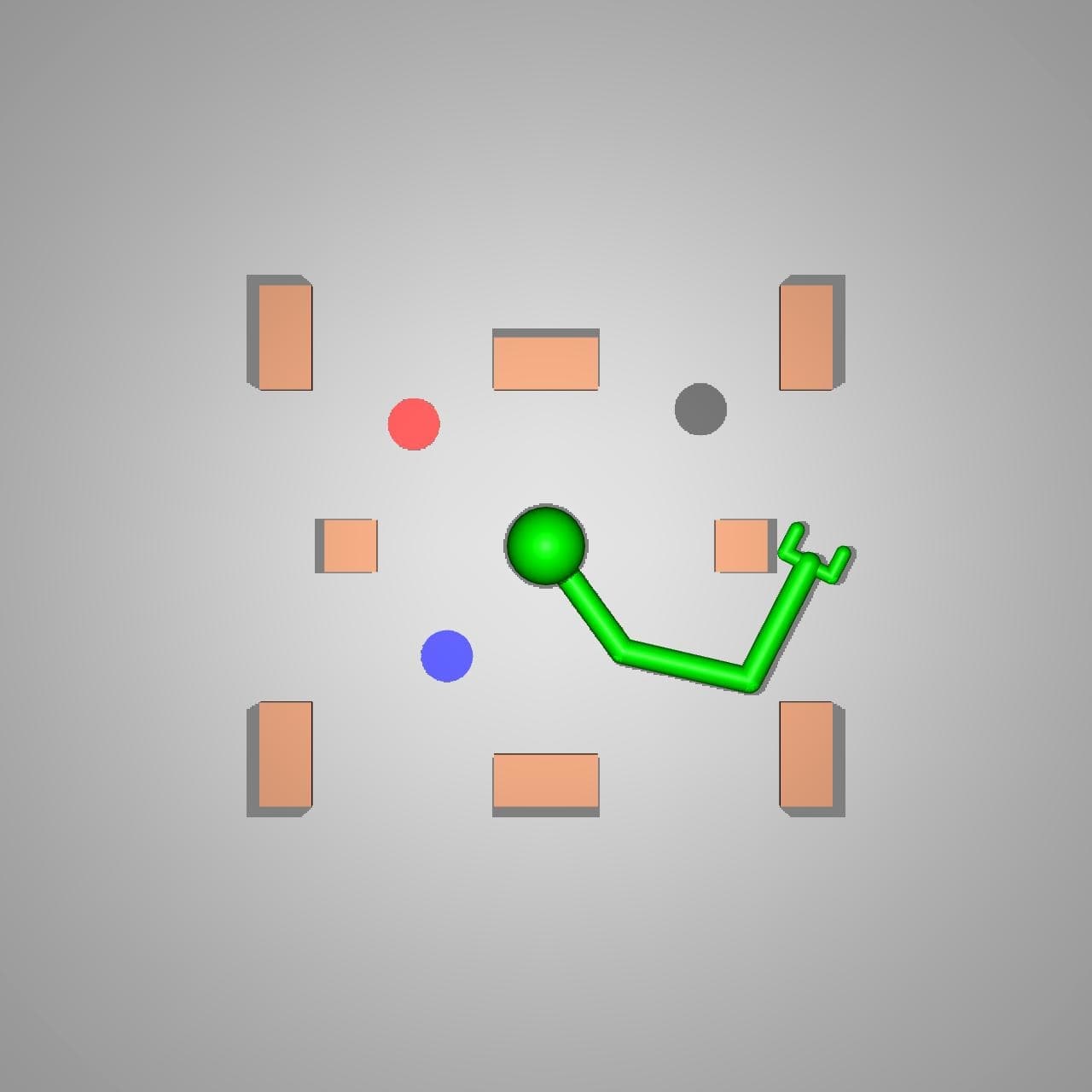}
    \caption{3 Link Reacher Robot around obstacles}
    \label{fig:reachertopdown}
\end{figure}

\subsection{Pusher}

\begin{figure}[H]
    \centering
    \includegraphics[width = 0.2 \textwidth]{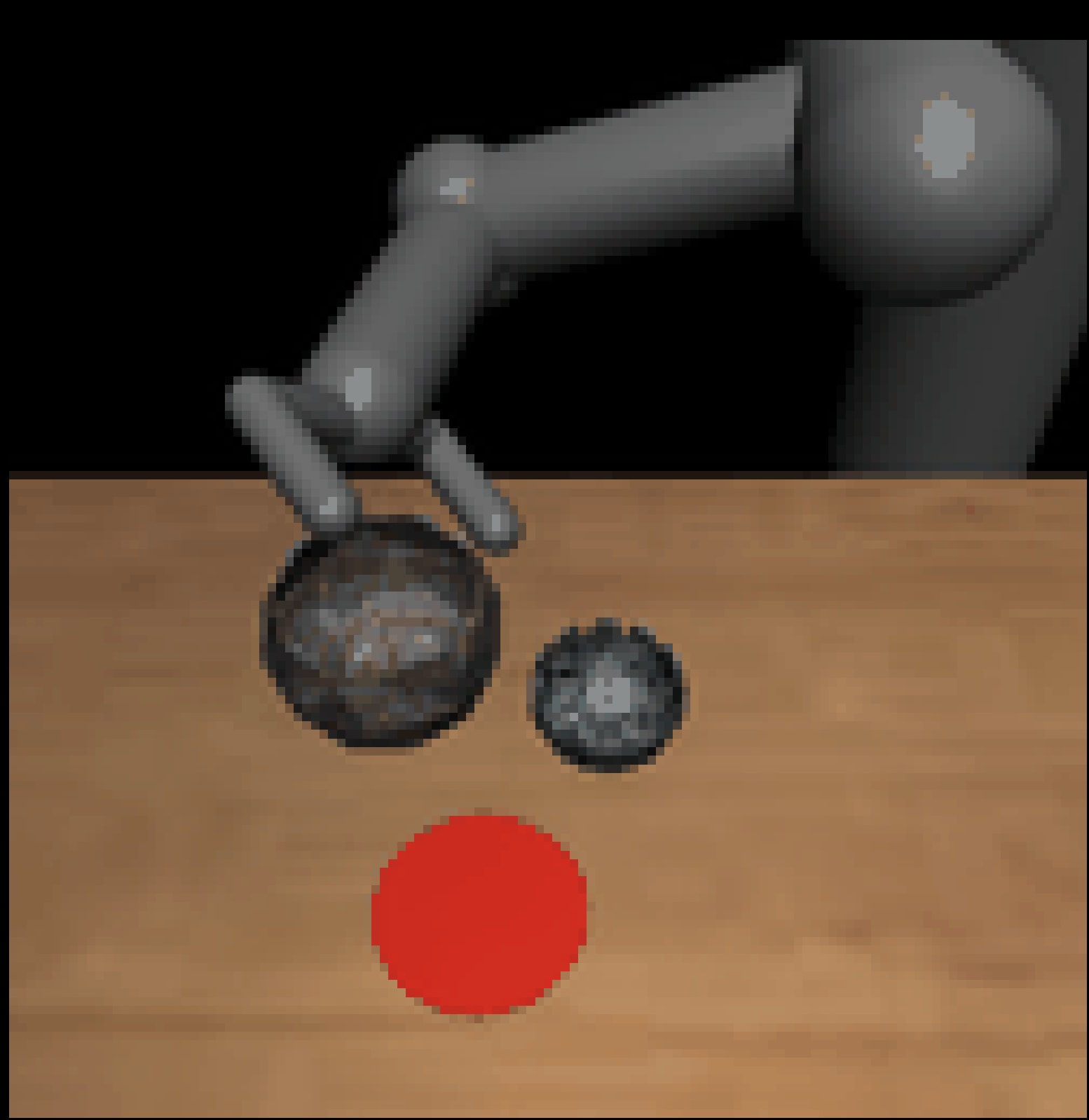}
    \caption{7-DoF  Pusher trying to push an object to a goal}
    \label{fig:pushertopdown}
\end{figure}



\section{Exponential of distance metrics for rewards}

As pointed out in subsection \ref{rlexplanation} of the main paper, we normalize our huber metrics that are derived out of the UPN representations through an exponential. This normalizes the reward functions we optimize for and in practice resulted in stable training with the PPO algorithm. We thank \href{https://xbpeng.github.io/}{Xue Bin (Jason) Peng} for communicating this trick to us. We adopted this transformation of the reward for all our reinforcement learning experiments. To be specific, our rewards are of the form $e^{-\sigma \textrm{Huber}(x_t, x_g)}$. The specific functional forms used for the different tasks are pointed out in Table \ref{sigmatable}.

\begin{table}[h]
\footnotesize
\caption{Environment Specific Details}
\label{sigmatable}
\vskip 0.15in
\begin{center}
\begin{small}
\begin{sc}
\begin{tabular}{lcccr}
\toprule
Environment & Reward\\
\midrule
nt (simple) & $e^{-2.5 \textrm{Huber}(x_t, x_g)}$ \\
5-link reacher  & $e^{ -\textrm{Huber}(x_t, x_g)}$  \\
Ant (hard) & $0.6 e^{-\textrm{Huber}(x_t, x_g)} + 0.4 e^{-2.5 \textrm{Huber}(x_t, x_g)}$  \\
Humanoid &   $0.6 e^{-\textrm{Huber}(x_t, x_g)} + 0.4 e^{-2.5 \textrm{Huber}(x_t, x_g)}$\\
Pushing  & $0.6 e^{-\textrm{Huber}(x_t, x_g)} + 0.4 e^{-2.5 \textrm{Huber}(x_t, x_g)}$\\

\bottomrule
\end{tabular}
\end{sc}
\end{small}
\end{center}
\vskip -0.1in
\end{table}

\section{Open Source Frameworks}
All our experiments were done using TensorFlow \cite{abadi2016tensorflow}, which allows automatic differentiation through the gradient updates during the inner loop planning. We also built our implementations of the tasks and reinforcement learning methods on top of OpenAI Gym~\cite{brockman2016openai} and OpenAI Baselines~\cite{baselines}. Our environments use the MuJoCo physics simulator~\cite{todorov2012mujoco} and its python bindings~\cite{mujocopy} developed by OpenAI. The AWS Docker setup was adapted from \url{https://github.com/anair13/selfsupervised/blob/master/Dockerfile}.

\end{document}